\documentclass[10pt]{article}
\usepackage{2022abstract}

\usepackage{amsmath}
\usepackage{amsfonts}
\usepackage{wrapfig}
\usepackage{xcolor}

\usepackage{sidecap}
\usepackage{moreverb,url}
\usepackage{multirow}

\title{From eye-blinks to state construction: diagnostic benchmarks for online representation learning} 

\author[1]{Banafsheh Rafiee}
\author[2]{Zaheer Abbas}
\author[1]{Sina Ghiassian}
\author[1]{Raksha Kumaraswamy}
\author[1,2]{Richard S. Sutton}
\author[3]{Elliot A. Ludvig}
\author[1,2]{Adam White}

\affil[1]{Department of Computing Science and the Alberta Machine Intelligence Institute (Amii), University of Alberta,
Edmonton, Canada}
\affil[2]{DeepMind Alberta, Edmonton, Canada}
\affil[3]{Department of Psychology, University of Warwick, Coventry, United Kingdom}

\begin{document} 
\maketitle

\section{Abstract} 
We present three new diagnostic prediction problems inspired by classical-conditioning experiments to facilitate research in online prediction learning.
Experiments in classical conditioning show that animals such as rabbits, pigeons, and dogs can make long temporal associations that enable multi-step prediction.
To replicate this remarkable ability, an agent must construct an internal state representation that summarizes its interaction history. 
Recurrent neural networks can automatically construct state and learn temporal associations.
However, the current training methods are prohibitively expensive for {\em online prediction}---continual learning on every time step---which is the focus of this paper.
Our proposed problems test the learning capabilities that animals readily exhibit and highlight the limitations of the current recurrent learning methods. 
While the proposed problems are nontrivial, they are still amenable to extensive testing and analysis in the small-compute regime, thereby enabling researchers to study issues in isolation, ultimately accelerating progress towards scalable online representation learning methods.

\noindent\Keywords{State construction}{classical conditioning}{diagnostic benchmarks}{reinforcement learning}\\

\noindent \corauthor{Banafsheh Rafiee}{rafiee@ualberta.com}

\section{Introduction}
\begin{figure}[h!]
\centering
  \includegraphics[width=0.4\linewidth]{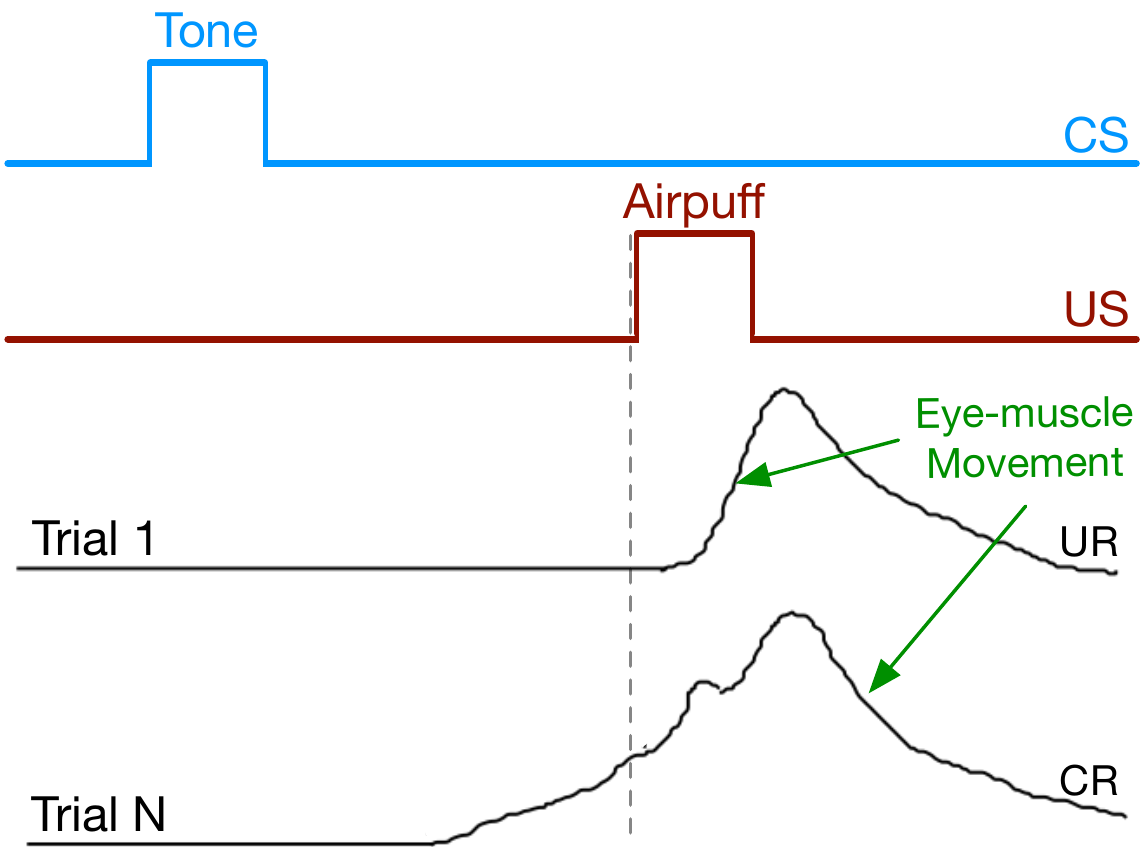}
  \caption{Eyeblink conditioning. After many pairings of the tone with the puff of air, the rabbit learns to close its inner eyelid (nictating membrane) before the puff of air is presented.}
  \label{fig:eyeblink}
\end{figure}

We consider the problem of multi-step prediction learning in a partially observable setting.
In the multi-step prediction learning problem, the agent's objective is to use its sensory experience to predict signals of interest multiple steps into the future, just like when a reinforcement learning agent must predict future reward. 
In the partially observable setting, the agent must also construct an internal representation that summarizes its experience, as the immediate sensory information may not be sufficient for making accurate long-term predictions. 
Consider, for example, a rabbit trained to preemptively close its eyes by predicting a puff of air using another predictive stimulus, such as a tone, as shown in Figure \ref{fig:eyeblink}. 
To appropriately time the eyeblink, the rabbit needs an internal representation of the elapsed time since the tone sounded.
Neural network solution methods can be used for such problems \citep{tallec2017unbiased,jaderberg2017decoupled,dehghani2018universal,gehring2017convolutional,nath2019training}. 
Researchers use a variety of benchmarks to evaluate the progress of the neural network solution methods -- toy problems, time-series data sets, NLP tasks, and large-scale navigation problems.
We focus on the case in which the agent learns \textit{online}: making and updating its predictions on every time step, even when the prediction target is not immediately available, as in temporal-difference (TD) learning \citep{sutton1988learning}.

Benchmarks in reinforcement learning are relevant for evaluating multi-step predictions, but most are based on the fully observable setting.
The Arcade Learning Environment (ALE) exhibits minor partial observability, but frame-stacking can be used to construct a state that can achieve good performance \citep{bellemare2013arcade,machado2018revisiting}. 
OpenAI-Gym \citep{brockman2016openai} and MuJoCo \citep{todorov2012mujoco} offer a wide variety of tasks inspired by problems in robotics that are partially observable when using only visual inputs. 
However, the focus is mostly on continuous actions and high-dimensional inputs from joint angles and velocities.
The DeepMind Lab contains several 3D simulation problems inspired by experiments in neuroscience \citep{beattie2016deepmind,wayne2018unsupervised}.
Researchers have used these problems to benchmark large-scale learning systems; unfortunately, such experiments require several billion steps of interaction and cloud-scale compute \citep{beattie2016deepmind,wayne2018unsupervised,parisotto2019stabilizing, fortunato2019generalization,espeholt2018impala}.

Diagnostic issue-oriented benchmarks serve different purposes than large-scale challenge problems. 
While the diagnostic benchmarks are simple, they still illuminate fundamental limitations of the existing methods.
For example, the eight-state Black and White problem highlights the need for tracking in partially observable problems \citep{sutton2007role}, and DeepSea highlights how dithering exploration can be arbitrarily inefficient even in a grid world ~\citep{osband2019deep}.
Such diagnostic problems isolate specific algorithmic issues, and progress on these problems represents progress on the specific issues.
Additionally, if a diagnostic benchmark has small compute requirements, then researchers can quickly evaluate new ideas and avoid the additional engineering complexity required to build high-performance, state-of-the-art architectures. 
Large problems often require complex architectures that can be difficult to analyze, and small implementation details can lead to incorrect conclusions \citep{engstrom2019implementation,tucker2018mirage}.
Robust statistical analysis, experiment repetition, and ablations can be challenging in large-scale benchmarks because of the excessive computational requirements (see \citet{machado2018revisiting,henderson2018deep,colas2018many}).

Inspired by animal learning, this paper contributes a set of diagnostic benchmarks for the partially observable online prediction problem.\footnote{The source code for our three benchmark problems is available \href{https://github.com/banafsheh-rafiee/From-Eye-blinks-to-State-Construction-Diagnostic-Benchmarks-for-Online-Representation-Learning}{here.}}
Our first problem, {\em trace conditioning}, requires an agent to predict a distal stimulus from a previously observed cue, just as a rabbit predicts an air puff based on a tone.
The challenge here is representational: how does the agent bridge the gap between the tone and the air puff in a way that is not specific to the particular arrangement or timing of the stimuli \citep{ludvig2012evaluating,sutton2018reinforcement}.
Our second problem, {\em noisy patterning}, is inspired by biconditional patterning experiments \citep{mackintosh1974psychology,harris2008negative}.
This problem tests the agent's ability to determine which observation signals to pay attention to, in the presence of noise and distracting stimuli.
Finally, our third benchmark, {\em trace patterning}, combines trace conditioning and noisy patterning and requires the agent to simultaneously discover the relevant observation signals and build their temporal representations.

Our second contribution is empirical.
We use the proposed diagnostic problems to conduct a comprehensive empirical study of several state-of-the-art recurrent learning architectures, including Long Short-Term Memory (LSTM) ~\citep{hochreiter1997long} and related Gated Recurrent Units (GRU)~\citep{cho2014properties}, trained via Truncated Back-prop Through Time (T-BPTT) \citep{williams1990efficient} and Real Time Recurrent Learning (RTRL)~\citep{williams1989learning}.
We systematically investigate each method's performance as we vary the key problem parameters.
We also introduce a simple input augmentation scheme based on memory traces, improving both T-BPTT and RTRL based methods.
In total, our results show that the proposed diagnostic problems can effectively isolate the limitations of the current training methods and help stimulate research in online representation learning.

\section{Related Work}
\label{sec:related_work}
In partially observable problems, the agent must construct an internal state to summarize the history of interaction in order to predict the future. This is often done by recurrent networks. An RNN uses hidden layers with recurrent connections trained via BPTT \citep{hopfield1982neural,elman1990finding}, in order to summarize the history of interaction. Storing network activations from the beginning of time is expensive, and so the update can be truncated $T$ steps back in time (i.e., T-BPTT) \citep{williams1990efficient}. 
This presents a trade-off. 
If the truncation window is short, the agent cannot learn long temporal dependencies. 
If the truncation window is long, however, the agent can learn long temporal associations, but computation and memory costs grow with $T$. If the truncation window is shortened, then most recurrent systems including basic RNNs and LSTMs (and GRUs) cannot learn temporal relationships longer than $T$ \citep{williams1990efficient}. This trade-off is particularly challenging in the online prediction setting where the agent's objective is to update and make a new prediction on each time step. Ideally, our state construction methods would be able to learn dependencies greater than $T$ without requiring proportional computation---as humans do.\footnote{Humans do not appear to perform more computations to remember further back in time, rather people appear to employ abstractions that lose precision the further back they remember.}

There are alternatives to T-BPTT, many based on RTRL; which is itself an approximation of the true gradient. 
For a fully connected network, RTRL requires quartic computation in the number of hidden states per step which makes online implementation with even modestly sized networks challenging \citep{williams1989learning}. 
Approximations of RTRL such as Unbiased Online Recurrent Optimization (UORO) \citep{tallec2017unbiased}, synthetic gradient methods \citep{jaderberg2017decoupled}, and SnAp \citep{menick2020practical} approximate the gradient back in time and thus suffer from the representability/computation tradeoff of T-BPTT. 
We did not include UORO and SnAp as baselines in our experiments; we instead included the results from RTRL which both these methods approximate. 
We showed that the performance of RTRL significantly deteriorates as the temporal associations become longer, suggesting that its recent approximations will also have difficulty with the proposed benchmarks. 
In addition, prior work \citep{nath2019training} found UORO to perform significantly worse than simpler T-BPTT variants in the related online predict k-steps ahead problem setting, suggesting that our benchmarks would be challenging for UORO.

Recent work has explored alternatives to overcome the trade-off, including alternative optimization schemes for RNNs~\citep{nath2019training}, and learned sparse attention mechanisms combined with feedforward networks~\citep{dehghani2018universal,gehring2017convolutional}. 
Fixed Point Propagation \citep{nath2019training} has not been extended to our discounted multi-step prediction setting (estimating value functions). 

Learned sparse attention mechanisms combined with feed-forward neural networks represent exciting alternatives for training RNNs. 
The best way to use attention strategies for partially observable reinforcement  learning is still evolving \citep{parisotto2019stabilizing,parisotto2021efficient,loynd2020working,chen2021decision,janner2021reinforcement}. 
\cite{chen2021decision} and \cite{janner2021reinforcement} use transformers in the offline reinforcement learning setting.
\cite{parisotto2019stabilizing} and \cite{parisotto2021efficient} stack long sequences of past observations in order to learn long temporal dependencies. 
Therefore, they require at least linearly more resources as the span of temporal dependencies increases, which reintroduces the truncation tradeoff.
Combining transformers with mini-batches skewed more towards recent experiences (as shown to be effective in RL \citep{zhang2017deeper}) represents an interesting next step. 
However, more work is required to extend it to our online multi-step prediction learning setting.
As these strategies are still beginning to be explored by the community, we leave these comparisons to future work.

Small diagnostic benchmarks like ours have a long history in online learning and reinforcement learning. Prior work on online supervised representation learning \citep{sutton2014online,mahmood2013representation}, step-size adaption methods \citep{sutton1992adapting,jacobsen2019meta}, and divergence in temporal difference learning \citep{baird1995residual,sutton2018reinforcement} all make use of small diagnostic test problems to evaluate progress. More generally, small issue-focused problems are used pervasively in reinforcement learning to isolate and study research questions (see \citet{sutton2018reinforcement}). 
The Deepmind Behavior Suite in many ways represents a modern attempt to organize and standardize a collection of interesting diagnostic test problems in reinforcement learning \citep{osband2019behaviour}, similar in spirit to the Reinforcement Learning Competitions of old \citep{whiteson2010report}.  
Recent work has shown that classic toy problems like Mountain Car and Acrobot can be used to highlight the advantages of fairly complex modern architectures like Rainbow \citep{ceron2021revisiting}, with a fraction of the computation typically required to run ALE experiments.
Our diagnostic benchmarks can be accurately thought of as Prediction Suite. 

\section{Classical Conditioning as Representation Learning}
\label{sec:AL}

The study of multi-step prediction learning in the face of partial observability dates back to the origins of classical conditioning. Pavlov was perhaps the first to observe that animals form predictive relationships between sensory cues while training dogs to associate the sound of a metronome with the presentation of food \citep{pavlov1927conditioned}. The animal uses the sound of a metronome (which is never associated with food in nature) to predict when the food will arrive, inducing a hardwired behavioral response. The ability of animals to learn the predictive relationship between stimuli is critical for survival. These responses could be preparatory like a dogs' salivation before food presentation or protective in case of anticipating danger like blinking to protect the eyes. Such predictions in the face of limited information are useful to humans too. You predict when the bus might stop next---and perhaps get off---based on the distal memory of the bell. You predict when the water from the tap might get too hot and move your hand in advance. The study of prediction, timing, and memory in natural systems remains of chief interest to those that wish to replicate it in artificial systems.

Some of the most relevant theories on multi-step prediction in animals have been explored in {\em trace conditioning}. In the classical setup, two stimuli are presented to the animal in sequence as shown in Figure \ref{fig:eyeblink}. The first is called the conditioned stimulus or CS (the predictive trigger) which usually takes the form of a light or tone. Then an unconditioned stimulus (US), such as a puff of air to the animal's eye, is presented which generates a behavioral response called the unconditioned response (UR)---the rabbit closes its inner eyelid. After enough pairings of the CS and US, the animal produces a conditioned response (e.g., closing the inner eyelid) after the CS---behaving in advance of the US. This arrangement is interesting because there is a gap, called the trace interval, between the offset of the CS and onset of the US where no stimuli are presented.
Empirically we can only reliably measure the strength and timing of the animal's anticipatory behavior: the muscles controlling the inner eyelid. However, the common view is that the rabbit is making a multi-step prediction of the US triggered by the onset of the CS that grows in strength closer to the onset of the US \citep{schneiderman1966interstimulus,sutton1990time,sutton2018reinforcement}, similar to the conditioned response in Figure \ref{fig:eyeblink}.

The mystery for both animal learning and Artificial Intelligence (AI) is how does the agent fill the gap? 
No stimuli occur during the gap and yet the prediction of the US rises on each time step. There must be some temporal generalization of the stimuli occurring inside the animal. Additionally, what is the form of the prediction being made, and what algorithm is used to update it? Previous work has suggested that the predictions resemble \textit{discounted returns} used in reinforcement learning \citep{dickinson1980contemporary,wagner1978expectancies}, sometimes called nexting predictions \citep{modayil2014multi}, which can be learned using temporal difference learning and eligibility traces (i.e., TD($\lambda$)). Indeed the TD-model of classical conditioning has been shown to emulate several phenomena observed in animals \citep{ludvig2012evaluating,ludvig2008stimulus,sutton1990time}.

On the question of representation or agent state, the answer is less clear. TD-models can generate predictions consistent with the animal data, but only if the state representation fills the gap between the CS and US in the right way \citep{ludvig2012evaluating,ludvig2009computational,williams2017intertrial}. A flag indicating the CS just happened, called the {\em presence representation}, will not induce predictions that increase over time, and a clock is not plausible given the range of timescales, the presence of other non-relevant distracting signals, and the massive number of predictive relationships an agent must learn in its lifetime \footnote{Ludvig's microstimulus representation can be viewed as a clock whose resolution gets worse over time \citep{ludvig2012evaluating}.}\citep{gallistel2011memory}. 
Hand-designed temporal representations do reproduce the animal data well \citep{ludvig2012evaluating,ludvig2008stimulus,ludvig2009computational,williams2017intertrial}, but their generality remains unclear. Ideally, the learning system could discover for itself how to represent different stimuli over-time in a way that (1) is useful across a variety of prediction tasks, and (2) requires computation and storage independent of the size of the trace interval. Animals do require more training to learn trace conditioning tasks with longer and longer trace intervals, but there is no evidence that the update mechanisms or representations fundamentally change as a function of the trace interval \citep{howard2013hippocampus}. 
Prior work by Rivest et al. has investigated an LSTM driven by temporal-difference errors as a model of cortical and dopaminergic neurons during trace conditioning \citep{rivest2014conditioning}, but Rivest's work did not focus on the impact of problem parameters like the trace interval on learning performance. The Rescorla-Wagner drift-diffusion model provides a reasonable account of trace conditioning \citep{luzardo2018rescorla}, but does not update predictions during the trial.

Trace conditioning represents a family of diagnostic problems with many potential variations. There could be several additional stimuli which are unrelated to the CS and US, called {\em distractors}. The CS and US could occur for different lengths of time and overlap in different ways. There can be multiple CSs and the US might only occur for particular ordering and configurations of the CSs. In {\em patterning} or {\em biconditional discrimination} experiments, for example, the CSs all occur at the same time step, but only a particular pattern of active and inactive CSs trigger the US (see \citet{harris2008negative}). 
Finally, we can combine these problems in a variety of ways mixing multi-step dependencies, distractors, and patterning. In this paper, we propose three variations as diagnostic benchmark problems for evaluating online multi-step prediction and state construction.

\section{From Animal Learning to Online Multi-step Prediction}
\label{sec:background}
We model our multi-step prediction task as an uncontrolled dynamical system. 
At every time step $t$, the agent observes stimuli ${\bf o}_t \in \mathbb{R}^d$, which includes $\text{CS}_t$ and $\text{US}_t$, and makes a prediction $v_t \in \mathbb{R}$ about the future value of the US.
The CS at time $t$ may be relevant to the prediction of the US in the future, and the observation ${\bf o}_t$ may contain distractors that are unrelated to the US---regardless ${\bf o}_t$ does not fully capture the current state of the system.
As discussed in Section \ref{sec:AL}, a suitable choice for formulating the US predictions is the expected discounted return or value function: $v_t \doteq \mathbb{E}[G_t | S_t]$ where 
\begin{equation}\label{eq:ret}G_t\doteq\sum_{k=0}^{\infty} \gamma^k \text{US}_{t+k+1}, \end{equation}
is called the {\em return} and $S_t$ is the unobserved state. 
$\gamma\in [0,1)$ is called the discount factor and defines the horizon of the prediction of the US. 

\begin{figure*}[]
\centering
  \includegraphics[width=0.75\linewidth]{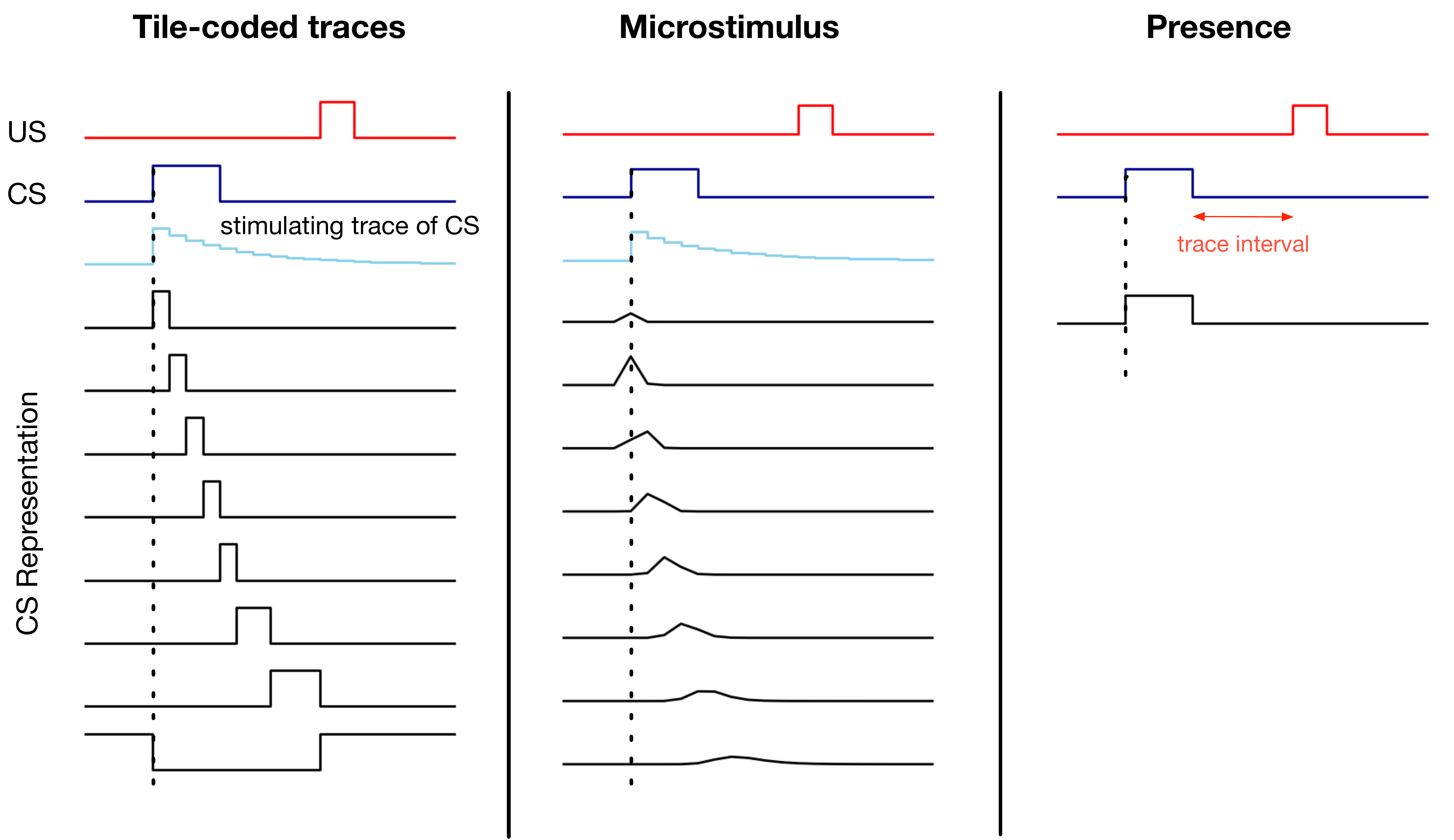}
  \caption{The stimulus representation for the tile-coded traces, microstimulus, and presence representations. The presence representation does not have any active features during the trace interval. This figure is adapted from \cite{ludvig2012evaluating}. }
   \label{fig:problem_1_rep}
\end{figure*}

\begin{figure}[h]
\centering
  \includegraphics[width=0.3\linewidth]{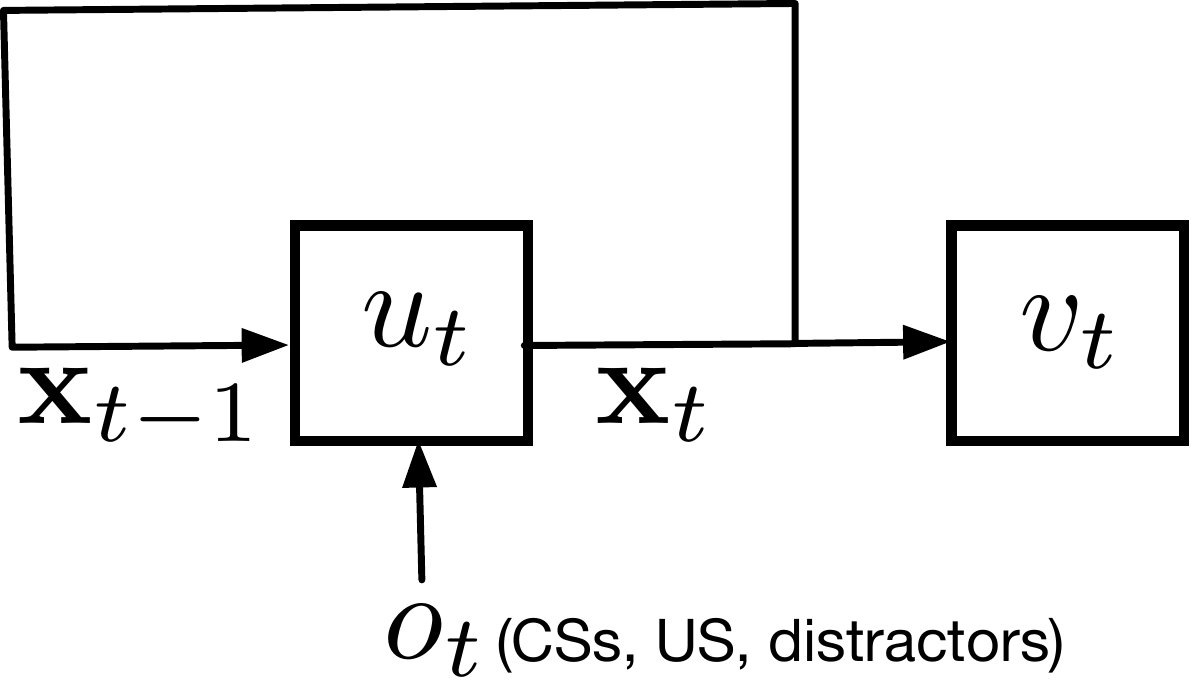}
  \caption{Recursive construction of ${\bf x}_t$ from $o_t$ and ${\bf x}_{t-1}$ where $o_t$ includes the CSs, the US, and the distractors, if any.}
   \label{fig:recurrence}
\end{figure}

We will incrementally estimate $v_t$ on each time step with semi-gradient temporal difference (TD) learning \citep{sutton1988learning}.
Semi-gradient TD is the most commonly used algorithm for these online prediction tasks and has appealing features relevant to our setting. TD is (1) simple and computationally frugal (linear complexity), and (2) efficient and accurate for learning multi-step predictions online from real data (see \citet{modayil2014multi}). Semi-gradient TD learns a parametric approximation $V_t \in \mathbb{R} \approx v_t$ by updating a vector of parameters ${\bf w}_t \in \mathbb{R}^d$ as follows:
\begin{align}
{\bf w}_{t+1} &\leftarrow {\bf w}_t + \alpha (\text{US}_{t+1} + \gamma V_{t+1} - V_t) {\bf z}_t \label{alg:td} \\
{\bf z}_t &\leftarrow \gamma \lambda {\bf z}_{t-1} + \nabla_{{\bf w}} V_t \nonumber
\end{align}
\noindent where $\alpha\in(0,1]$ is the learning rate and $\lambda\in[0,1]$ controls the decay of eligibility trace ${\bf z}_t \in \mathbb{R}^d$. The precise form of $V_t$ depends on the parameterization scheme. In the linear case, $V_t \doteq {\bf x}_t^T{\bf w}_t$ and $\nabla_{w} V_t =   {\bf x}_t$, where ${\bf x}_t \in \mathbb{R}^d$ is a vector of features constructed from ${\bf o}_t$. In the non-linear case, $V_t$ can be computed by a neural network and $\nabla_{w} V_t$ by backpropagation. More generally, $\text{US}_{t+1}$ can be any component of ${\bf o}_t$ in Eq. \ref{eq:ret} and \ref{alg:td} allowing prediction of any component of the observations (as in \citet{modayil2014multi}). 

In this paper, we investigate different approaches to constructing ${\bf x}_t$. 
One approach is to simply form an exponentially-weighted decaying memory of each component of ${\bf o}_t$, or {\em stimulating trace},\footnote{A stimulating trace of the observation is different from the eligibility trace vector ${\bf z}$. ${\bf z}$ is part of the update mechanism and does not impact the representational capacity of ${\bf x}$. Mozer was the first to investigate stimulating traces as input to neural network representation learning \citep{mozer1989focused}.} and then apply a non-linear mapping to produce ${\bf x}_t$. 
Each component of stimulating trace, ${\bf y}_{t}$, corresponds to one component of ${\bf o}_t$ and is set to $1$ at the onset of the corresponding observation and decays immediately after the observation onset following $y_{t+1} = \tau y_t$ where $0<\tau<1$ is the decay parameter. 
Our tile-coded traces representation applies tile coding\footnote{See \citep{sutton2018reinforcement} for an in-depth treatment of tile coding.} to the stimulating traces of ${\bf o}_t$. 
In this case, the quality of ${\bf x}_t$ depends on both the tile coding parameters and the exponential decay rate of the stimulating trace. 
The so-called microstimulus representation, used in prior computational modeling of trace conditioning, is also a fixed feature construction approach dependent on hyper-parameters set by the designer. 
The microstimulus is formed from a set of overlapping Gaussian basis functions with the heights forming an exponential decay of ${\bf o}_t$ achieved by using larger standard deviations for each gaussian \citep{ludvig2012evaluating,ludvig2008stimulus,hull1939problem}.
Figure \ref{fig:problem_1_rep} shows an example of the stimulating trace of the CS and how the representation constructed by tile-coded traces ($1$ tiling $8$ tiles) and microstimulus (8 Guassians) for the CS unfold over time. 
 
Alternatively, ${\bf x}_t$ can be constructed recursively from ${\bf o}_t$ and ${\bf x}_{t-1}$ using a non-linear state update function ${\bf x}_t \doteq u({\bf x}_{t-1}, {\bf o}_t)$.
See Figure \ref{fig:recurrence}. 
The tile-coded traces and microstimulus representations represent particular instantiations of $u$ that never change during learning. 
We can also think of $u$ constructing ${\bf x}_t$ as a recurrent neural network. We consider both the case where $u$ is fixed at the beginning of learning, also known as echo state networks  \citep{jaeger2001echo}, as well as the case where T-BPTT or RTRL changes $u$ on each time step. 
In the case of echo state network, there are three groups of incoming weights to the hidden layer: 1) the input weights from the input to the hidden layer 2) the internal weights from the hidden layer to itself and 3) the feedback weights from the output layer to the hidden layer.
All the incoming weights to the hidden layer are fixed at the beginning of learning and only the weights from the features to the output are learned. 
In contrast, in the case of learning with T-BPTT and RTRL not only the agent's predictions of $v_t$ are updated, but also the function $u_t$ is learned.

Using T-BPTT and RTRL to train RNNs and their variants in an online setting is not new, nor is the application of such architectures to multi-step TD prediction targets. 
We followed standard practice in implementing these methods. 
For T-BPTT with truncation length $T$, when making an update at time $t$, we unroll the RNN for $T$ steps. We set the initial hidden state to ${\bf x}_{t-T-1}$. 
Then we compute the hidden states and the value predictions along the observation sequence ${\bf o}_{t-T}, ..., {\bf o}_{t-1}$. 
After computing the value predictions $V_{t-T}$ to $V_{t-1}$, we use them as a mini-batch to update the parameters of the network using backpropagation.

For RTRL, on the other hand, we update the parameters throughout the training sequence on every time step, while still carrying forward a stale Jacobian that tracks sensitivity to the old parameters (See \cite{menick2020practical}). 
  
  

\section{Trace Conditioning: Learning to Fill the Gap}
\label{sec:test_problem_1}
\begin{figure}[h!]
\centering
  \includegraphics[width=0.4\linewidth]{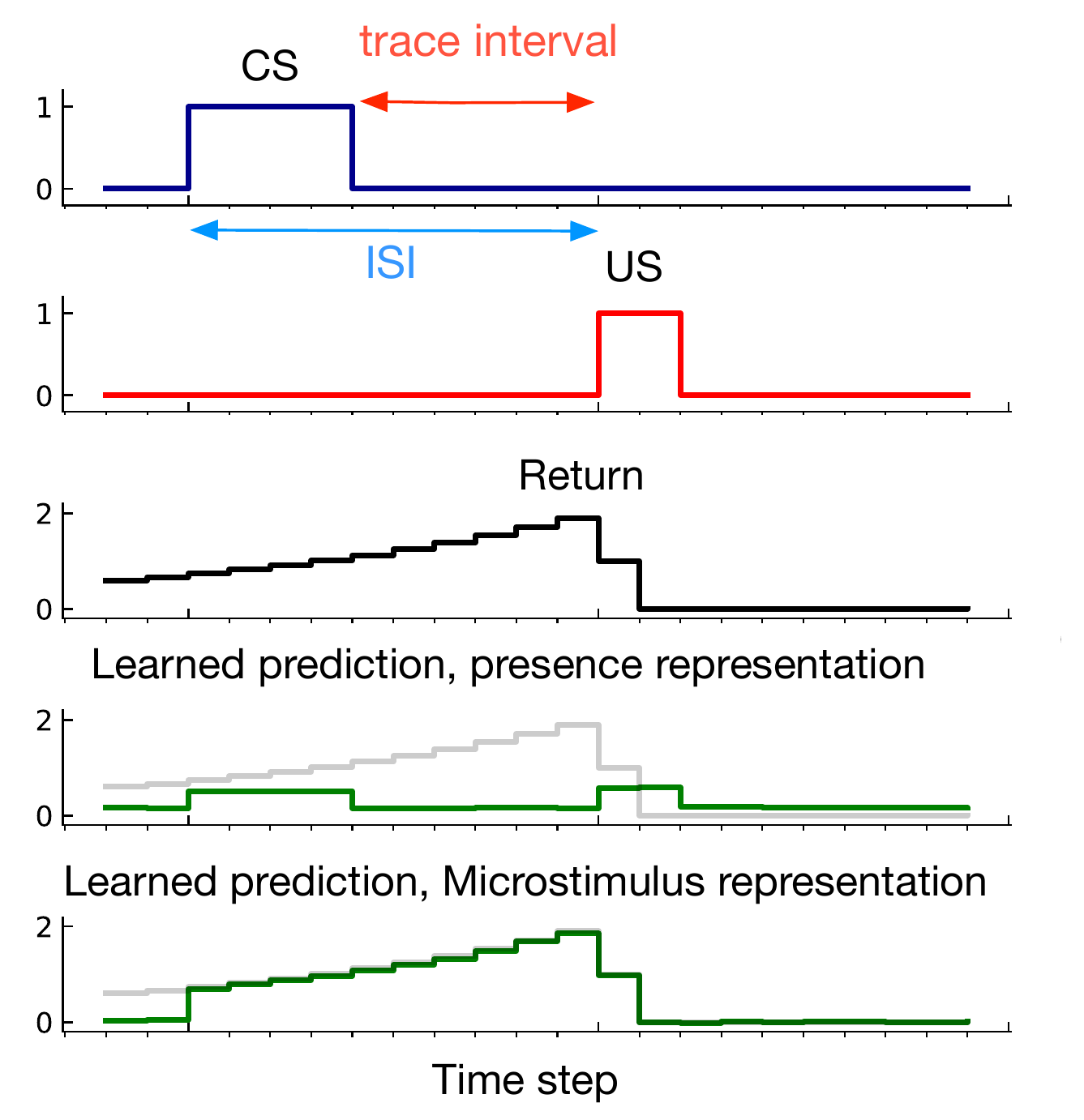}
  \caption{An example of learned predictions in {\bf trace conditioning}. The {\em return} defined in Eq. \ref{eq:ret} is the target of prediction. 
  Rows 4 and 5 show predictions using the presence and microstimulus representations after 200,000 time steps learning. 
  Microstimulus successfully predicted the US---matching the return---the presence representation failed to predict the US.
  The predictions never go to zero like the return because all representations use a bias feature and even after 200,000 steps the predictions continue to update.}
   \label{fig:problem_1_signals}
\end{figure}

Our first diagnostic problem, \textit{trace conditioning}, is inspired by classical conditioning experiments described in Section \ref{sec:AL}. 
The problem is made up of a series of trials in each of which a sequence of stimuli are presented: the CS followed by the US. 
On each trial, the CS lasts for $4$ time steps, and is followed by a long gap and then the US which lasts for $2$ time steps. 
The time from the CS onset to the US onset is called the {\em inter-stimulus interval} (ISI). 
In this problem, the ISI is drawn from a uniform distribution.
The time from the US onset to the start of the next trial is called the {\em inter-trial interval} (ITI). 
The ITI is uniformly sampled from $(80,120)$. 
$\gamma$ is set according to the ISI:  $\gamma = 1 - \frac{1}{\mathbb{E}(\text{ISI})}$.
This allows the time horizon of the return to match the ISI. 
Figure \ref{fig:problem_1_signals} provides an example trial including the CS, US, and return for a case where ISI $\sim \text{Unif}(7,13)$.

We also include several binary distractor stimuli that do not contain any information about the US. 
The distractors are drawn from a Poisson distribution with different frequencies and each lasts for $4$ time steps. 
The frequency varies from distractor to distractor.
One distractor occurs on average every $10$ steps, another every $20$ steps, and so on, up to one distractor that occurs every $100$ steps on average.
Note that they also occur during the ITI.

To understand why this problem could be challenging for a learning system, consider learning to predict using the presence representation. 
This representation contains one binary feature per stimulus which is activated only when the corresponding stimulus is present. 
The presence feature corresponding to the CS is active during the CS activation as shown in Figure \ref{fig:problem_1_rep}. 
However, during the trace interval, between the offset of the CS and the onset of the US, no feature is active (only the bias feature, which has a small weight associated with it is active) and therefore, the trace interval is not represented by the presence representation. 
As a result, as shown in Figure {\ref{fig:problem_1_signals}, the presence representation has a close to zero prediction during the trace interval.

To understand what a good prediction looks like, consider the microstimulus and tile-coded traces representations.
During the empty gap between the CS offset and the US onset, the microstimulus and tile-coded traces representations have active features constructed from a trace of the CS (see Figure \ref{fig:problem_1_rep}). 
As a result, they successfully associate the CS with the US (see the predictions for the microstimulus representation in Figure \ref{fig:problem_1_signals}).  
Note that the return reaches its maximum just before the US onset and steps downward after. 
This happens because the discounted sum of future USs is maximal just before the US onset: at this instant in time the US is multiplied by the largest possible values of the discount factor, $\gamma$. 
This temporal profile is consistent with previous work on {\em Nexting} \citep{modayil2014multi} and computational modeling in animal learning \citep{ludvig2012evaluating}.

Note that the prediction increases only after the CS onset whereas the return has non-zero values before the CS onset. 
This makes sense because there is a significant time between each trial and thus the onset of the CS is unpredictable by design---just like in trace conditioning experiments with animals.

%

In the trace conditioning benchmark, we experimented with two groups of representations as baselines.
The first group includes fixed representations: microstimulus, tile-coded traces, and echo state network (See Section \ref{sec:background} for the explanation about these representations). 
Microstimulus and tile-coded traces are expert-designed representations and include a bias feature that is always $1$. 
We adjusted the stimulating trace decay parameter for microstimulus and tile-coded traces according to the ISI: $\frac{1}{\mathbb{E}(\text{ISI})}$.
For echo state networks, all the three sets of weights contributing to constructing the hidden state were initialized and fixed at the beginning of learning. 
The input weights and feedback weights were initialized using a binomial distribution and scaled by an input scaling parameter. 
The internal weights were initialized in such a way that the spectral radius of the corresponding matrix is less than $1$ and its density is small.

The second group of representations includes those learned by recurrent neural networks: Vanilla-RNN, LSTM, and GRU. We used Haiku library for implementing the Vanilla-RNN, LSTM, and GRU architectures. We evaluated both T-BPTT and RTRL for computing the gradient of the value function with respect to the network's weights. 

\begin{table*}
\centering
\footnotesize 
\begin{tabular}{|c|c|c|c|c|c|c|c|c|}
\hline
\textbf{Problem}                                                                                              & \textbf{\begin{tabular}[c]{@{}c@{}}Representation \\ Method\end{tabular}} & \textbf{\begin{tabular}[c]{@{}c@{}}Number of \\ Tiles/RBFs\end{tabular}} & \textbf{\begin{tabular}[c]{@{}c@{}}Hidden \\ Layer\\ Size\end{tabular}} & \textbf{\begin{tabular}[c]{@{}c@{}}Spectral \\ Radius\end{tabular}} & \textbf{\begin{tabular}[c]{@{}c@{}}input \\ scaling\end{tabular}} & \textbf{\begin{tabular}[c]{@{}c@{}}W\_h \\ density\end{tabular}} & \textbf{\begin{tabular}[c]{@{}c@{}}Truncation \\ Length\end{tabular}} & \textbf{Step-size}                                                                             \\ \hline
\multirow{7}{*}{\textbf{\begin{tabular}[c]{@{}c@{}}Trace Conditioning\\ and\\ Trace Patterning\end{tabular}}} & Presence                                                                  & -                                                                        & -                                                                          & -                                                                   & -                                                                 & -                                                                & -                                                                     & \multirow{12}{*}{\begin{tabular}[c]{@{}c@{}}$3\times10^{-6}$, $10^{-5}$, \\$3\times10^{-5}$, $10^{-4}$, \\ $3\times10^{-4}$, $10^{-3}$\end{tabular}} \\ \cline{2-8}
                                                                                                              & Microstimulus                                                             & 4, 8, 16, 32                                                             & -                                                                          & -                                                                   & -                                                                 & -                                                                & -                                                                     &                                                                                                \\ \cline{2-8}
                                                                                                              & Tile-coded-traces                                                         & 2, 4, 8, 16                                                              & -                                                                          & -                                                                   & -                                                                 & -                                                                & -                                                                     &                                                                                                \\ \cline{2-8}
                                                                                                              & Vanilla-RNN                                                               & -                                                                        & 10, 20, 40                                                                 & -                                                                   & -                                                                 & -                                                                & 5, 10, 20, 40                                                         &                                                                                                \\ \cline{2-8}
                                                                                                              & GRU                                                                       & -                                                                        & 10, 20, 40                                                                 & -                                                                   & -                                                                 & -                                                                & 5, 10, 20, 40                                                         &                                                                                                \\ \cline{2-8}
                                                                                                              & LSTM                                                                      & -                                                                        & 10, 20, 40                                                                 & -                                                                   & -                                                                 & -                                                                & 5, 10, 20, 40                                                         &                                                                                                \\ \cline{2-8}
                                                                                                              & ESN                                                                       & -                                                                        & 100, 1000                                                                          & 0.9, 0.99, 0.999                                                    & 0.1, 0.5                                                          & 0.05, 0.1                                                        & -                                                                     &                                                                                                \\ \cline{1-8}
\multirow{5}{*}{\textbf{Noisy Patterning}}                                                                    & Presence                                                                  & -                                                                        & -                                                                          & -                                                                   & -                                                                 & -                                                                & -                                                                     &                                                                                                \\ \cline{2-8}
                                                                                                              & Vanilla-RNN                                                               & -                                                                        & 10, 20, 40                                                                 & -                                                                   & -                                                                 & -                                                                & 5                                                                     &                                                                                                \\ \cline{2-8}
                                                                                                              & GRU                                                                       & -                                                                        & 10, 20, 40                                                                 & -                                                                   & -                                                                 & -                                                                & 5                                                                     &                                                                                                \\ \cline{2-8}
                                                                                                              & LSTM                                                                      & -                                                                        & 10, 20, 40                                                                 & -                                                                   & -                                                                 & -                                                                & 5                                                                     &                                                                                                \\ \cline{2-8}
                                                                                                              & ESN                                                                       & -                                                                        & 100, 1000                                                                          & 0.9, 0.99, 0.999                                                    & 0.1, 0.5                                                          & 0.05, 0.1                                                        & -                                                                     &                                                                                                \\ \hline
\end{tabular}
\caption{Parameter sweeps for the three benchmarks.}
\label{tab:param}
\end{table*}

\begin{figure*}[h!]
\centering
  \includegraphics[width=0.8\linewidth]{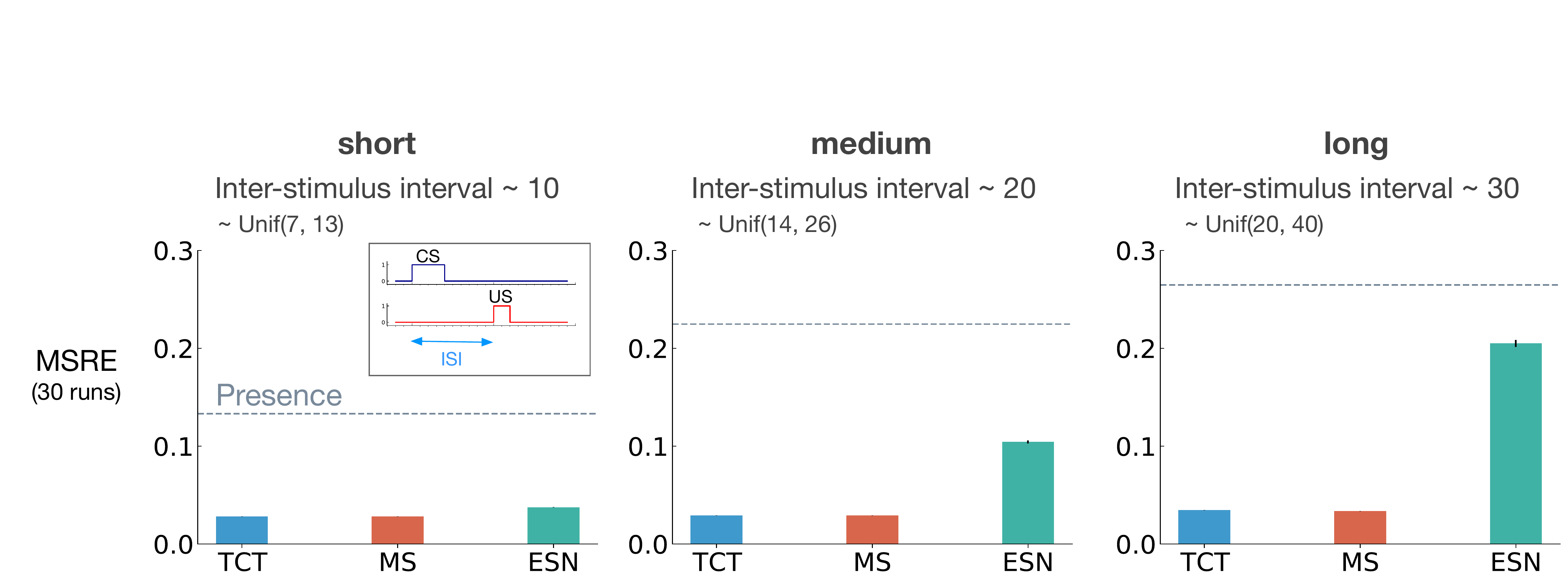}
  \caption{The interaction between ISI and truncation level in {\bf trace conditioning} for fixed representations: tile-coded traces (TCT), microstimulus (MS), and echo state network (ESN). 
  Each subplot corresponds to one setting of short, medium, and long ISI. 
  A mini picture of the CS and US timings is included in the leftmost subplot. 
  The y-axis is the MSRE. Lower is better. 
  The results are calculated over 2 million steps and averaged over 30 runs. 
  (Standard error bars are plotted but in some cases are not visible due to being small). 
  The error level for the presence representation is plotted in each subplot as a dotted line for comparison.
  In the short setting, all methods performed well. 
  Microstimulus and tile-coded traces performed well across all settings. 
  The performance of echo state network, however, deteriorated as ISI got larger.
  }
   \label{fig:domain1-fixed}
\end{figure*}


For each of these representations, we used semi-gradient TD$(\lambda)$ and ADAM optimizer with $\beta_1 = 0.9$, $\beta_2 = 0.999$, and $\epsilon = 10^{-8}$ \citep{kingma2014adam}.
For the fixed representations, we set the eligibility traces parameter, $\lambda$, to $0.9$. For the RNNs, we used $\lambda = 0$.

To evaluate the performance, we computed the Squared Return Error (SRE): $(\hat{v}(S_t, w_t) - G_t)^2$. We then averaged the SRE over all time steps resulting in a Mean Squared Return Error (MSRE). 
We studied the effect of the ISI on the performance of the baseline representation methods considering three cases: 1) short: ISI $\sim \text{Unif}(7, 13)$, 2) medium: ISI $\sim \text{Unif}(14, 26)$, and 3) long: ISI $\sim \text{Unif}(20, 40)$, with expected ISI equal to $10,20,30$ for the 3 settings respectively.

We swept over the parameters of each representation method. 
See Table \ref{tab:param}. 
The parameter sweeps included the step-size for all the methods, the number of Tile/RBFs for tile-coded-traces/microstimulus, the hidden layer size for the RNNs and echo state network, and the spectral radius, input scaling, and internal connections density for the echo state network.
For tile-coded traces, we used $2$ tilings and for microstimulus, we set the standard deviation of the RBFs to $0.8$.
For RNNs trained with T-BPTT, we swept over T-BPTT truncation length.
For all RNNs, the number of hidden layers was set to one. 

We ran each method with each of its parameter settings for $5$ runs and $2$ million time steps. 
We then computed MSRE averaged over the $5$ runs and selected the parameter setting that resulted in the lowest level of MSRE. 
After optimizing the parameters, we ran each method with its best parameter setting for $30$ runs and averaged the result. 
We calculated standard errors for each method to measure how far the sample means are from the true population means. 
We then plotted the MSRE averaged over $30$ runs and standard error bars with non overlapping standard error bars for two methods suggesting significant difference in their performance.


Figure \ref{fig:domain1-fixed} shows MSRE for fixed representations for short, medium, and long ISI. 
The y-axis is MSRE averaged over 30 runs.
The level of error for the presence representation is shown with a dotted grey line for comparison.

The expert designed fixed representations of microstimulus and tile-coded traces performed well across all ISI settings; however, echo state network failed to capture longer temporal dependencies. 
In the short setting, all fixed representations performed well. 
As ISI got larger, echo state network performed worse and approached the level of error of the presence representation.
This is likely due to the fact that echo state networks trade-off prediction accuracy for computation. 

Figure \ref{fig:domain1-rnn} shows MSRE for representations learned by T-BPTT and RTRL for short, medium, and long ISI. 
In each subplot, multiple bars are shown for each of Vanilla RNN, LSTM, and GRU architectures. For each architecture, the four left bars correspond to T-BPTT with $T=5$, $T=10$, $T=20$, and $T=40$. The right bar corresponds to the result for RTRL.  

\begin{figure*}[h!]
\centering
  \includegraphics[width=0.8\linewidth]{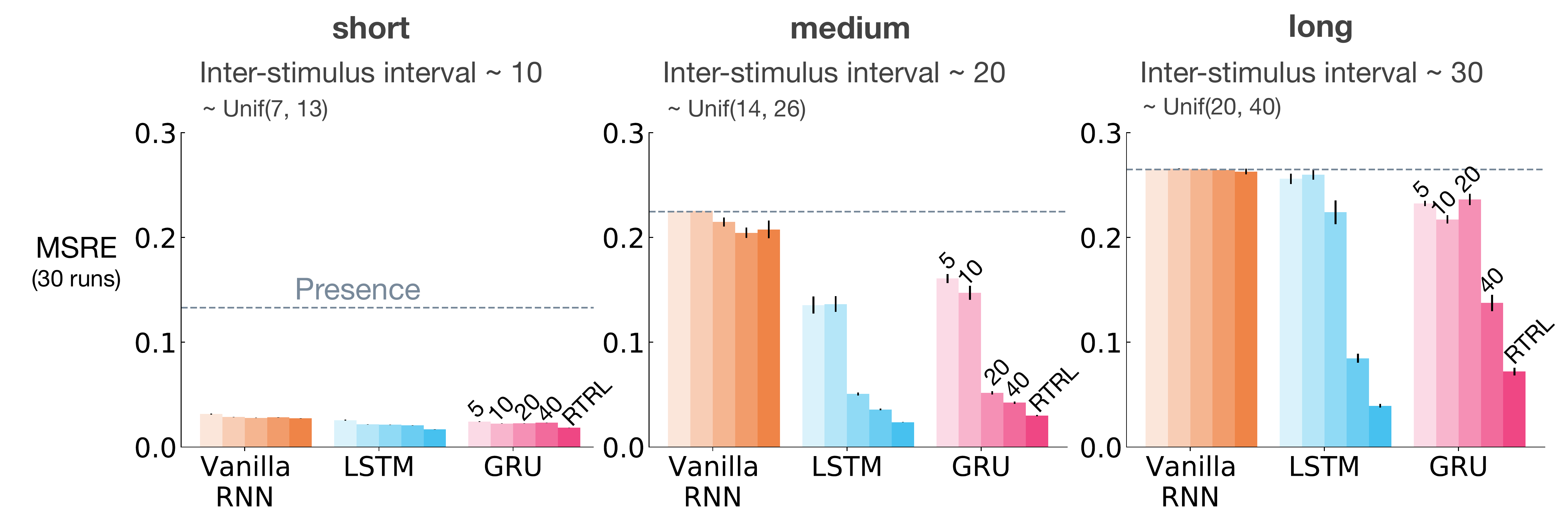}
  \caption{The interaction between ISI and truncation level in {\bf trace conditioning} for representations learned by T-BPTT and RTRL. 
  Each subplot corresponds to one setting of ISI. 
  In each subplot, multiple bars are plotted for Vanilla RNN, LSTM, and GRU. 
  For each architecture, the left four bars correspond to T-BPTT with different truncation levels and the right bar corresponds to RTRL. 
  The y-axis is the MSRE with lower better. 
  The results are calculated over 2 million steps and averaged over 30 runs. 
  Standard error bars are included in the plot. 
  With short ISI all methods performed well and the T-BPTT based methods worked with all $T$'s. 
  In the medium setting, we see basic RNNs performed poorly, and LSTMs and GRUs required truncation at or greater than expected ISI (20) to perform well. 
  In the long setting, we see that none of the T-BPTT based methods performed well, even with $T$ greater than expected ISI. 
  Across all three problem settings, RTRL-based LSTMs learned accurate predictions.
  }
   \label{fig:domain1-rnn}
\end{figure*}

In the short setting, the representations learned by both T-BPTT and RTRL performed well for all architectures, reaching a much lower level of error compared to the presence representation.

RNNs trained with T-BPTT were sensitive to the length of the truncation window, and the sensitivity became more pronounced as ISI got larger (Figure \ref{fig:domain1-rnn}).
To better understand this, let us contrast the performance of T-BPTT with that of the RTRL variants, which are roughly equivalent to T-BPTT for  $T=\infty$ 
(since when $T=\infty$, T-BPTT computes the gradient all the way back in time, resulting in a gradient roughly the same as the one computed by RTRL). 
In the medium setting, the T-BPTT variants for LSTMs and GRUs performed similarly to the RTRL counterparts only when the truncation window was greater than or equal to 20 -- the expected ISI (Figure \ref{fig:domain1-rnn}, middle column).
This effect was even stronger in the long setting (Figure \ref{fig:domain1-rnn}, right column). 
This result is one example of the efficacy of trace conditioning as a diagnostic benchmark -- it clearly isolates the trade-off introduced by the T-BPTT algorithm. 

There was a significant drop in the performance of Vanilla RNNs as we increased the expected ISI and larger truncation window did not help improve performance much.
This is likely due to the vanishing gradient problem \citep{hochreiter2001gradient}.
Vanilla RNN trained with RTRL also failed to capture longer dependencies.
This is in contrast to the LSTM and GRU variants trained with RTRL.

Our results suggest that further algorithmic improvements are required for solving the trace conditioning problem. 
While the expert designed fixed representations perform robustly across all ISI settings, they do not automatically discover useful features, and thus are not scalable.  
RTRL also performs well in all cases;  however, it is not computationally feasible. 
Finally, T-BPTT's performance is highly sensitive to the truncation parameter, requiring much more computation for learning longer temporal dependencies. 
Later we will discuss a simple algorithm that we tried to improve performance.





\section{Noisy Patterning}
\label{section:Test problem 2}
The trace conditioning benchmark is an idealization because there is only one signal of interest: the CS. The agent need not figure out which parts of its input stream to focus on---it is purely a temporal memory problem. Our second diagnostic benchmark, \textit{noisy patterning}, does not make this assumption. In noisy patterning, the agent must predict a binary outcome which only occurs if a particular pattern of stimuli is presented \citep{mackintosh1974psychology,harris2008negative}. 
To do so, it has to both figure out which parts of the input to pay attention to in the presence of noise and distractors and also make nonlinear features to identify the patterns of interest.
This is similar to the ``Blooming, Buzzing Confusion" visual stream that infants experience---they must learn what to pay attention to and ignore \citep{james2007principles}. 
In robot terms, the equivalent would be which sensors the agent should pay attention to, to avoid damage or gain additional reward. Similar problems have been studied in supervised learning \citep{sutton1992adapting,sutton2014online, mahmood2013representation}.

\begin{figure}[t]
\centering
  \includegraphics[width=0.5\linewidth]{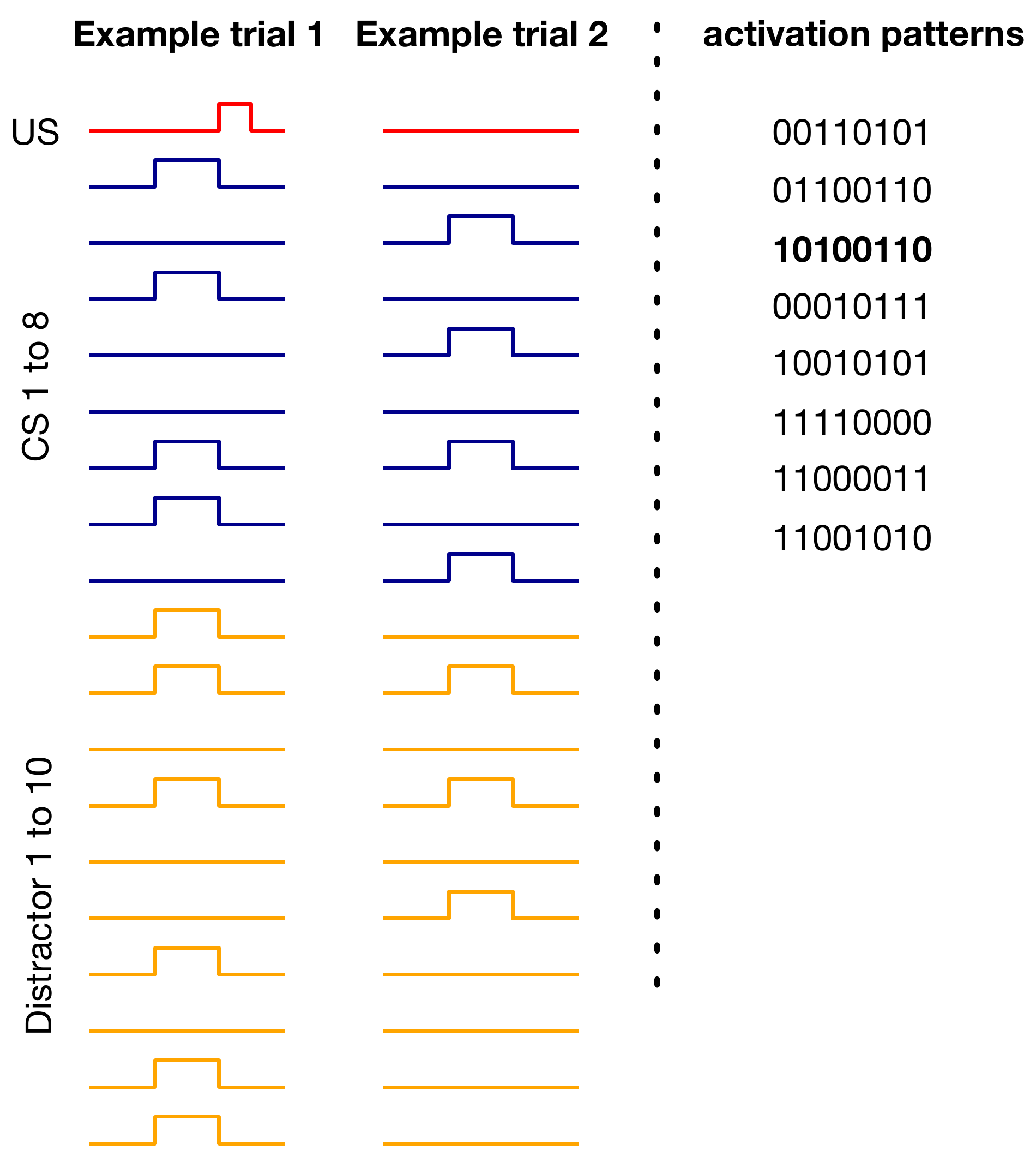}
  \caption{Example trials for noisy patterning in the case of $8$ CSs, $8$ activation patterns, $10$ distractors, and $10$ percent noise. {\tt10100110} is one of the $8$ activation patterns. In the example trial on the left, the pattern of the CSs matches this pattern and the US gets activated as a result. In the example trial on the right, however, the pattern of the CSs does not match any of the activation patterns resulting in US remaining $0$.}
   \label{fig:problem_2_signals}
\end{figure}

Noisy patterning is analogous to positive/negative patterning in psychology. It considers a situation where non-linear combinations of CSs activate the US. 
As we discussed in Section \ref{sec:AL}, in negative patterning each CS in isolation activates the US but their combination does not. 
Interestingly, these problems correspond to famous logical operations like XOR, which are famously unsolvable by single-layer neural networks.
While neural networks with more than one layer can easily learn patterning problems like XOR, some of the approaches considered in this paper, such as microstimulus, fail to solve them. 
To make the benchmark more challenging we designed the benchmark such that multiple configurations of the CSs activate the US and added distractors and noise.

This benchmark includes $n$ CSs and one US. 
There are $k$ configurations of the CSs that activate the US. 
We refer to these configurations as activation patterns. 
Each trial starts with the CSs getting a value of $0$ or $1$. 
If the value of the CSs matches an activation pattern, the US becomes $1$ in $4$ time steps (i.e. ISI equals 4). (In contrast to trace conditioning, the ISI is fixed.)
The ITI is uniformly sampled from $(80,120)$. 
We designed the benchmark such that in half of the trials, one of the activation patterns occurs each of which includes $n/2$ activated CSs and $n/2$ non-activated CSs. 
The benchmark also includes $m$ distractors, which occur at the same time as the CSs but do not contribute to the US activation. 
We also add noise such that in $x$ percent of the trials, an activation pattern occurs but the US remains $0$ or a non-activating pattern occurs and the US gets activated. 
$\gamma$ is set to $1 - \frac{1}{\text{ISI}}=0.75$. 
Two example trials for a case with $8$ CSs, $8$ activation patterns, $10$ distractors, and $10$ percent noise are shown in Figure \ref{fig:problem_2_signals}. In the example on the left, the pattern of the CSs matches one of the $8$ activation patterns. Therefore, the US gets activated. In the example on the left, however, the pattern of the CSs does not match any of the activation patterns. As a result, the US remains $0$. 


\begin{figure*}[]
  \centering
  \includegraphics[width=1\linewidth]{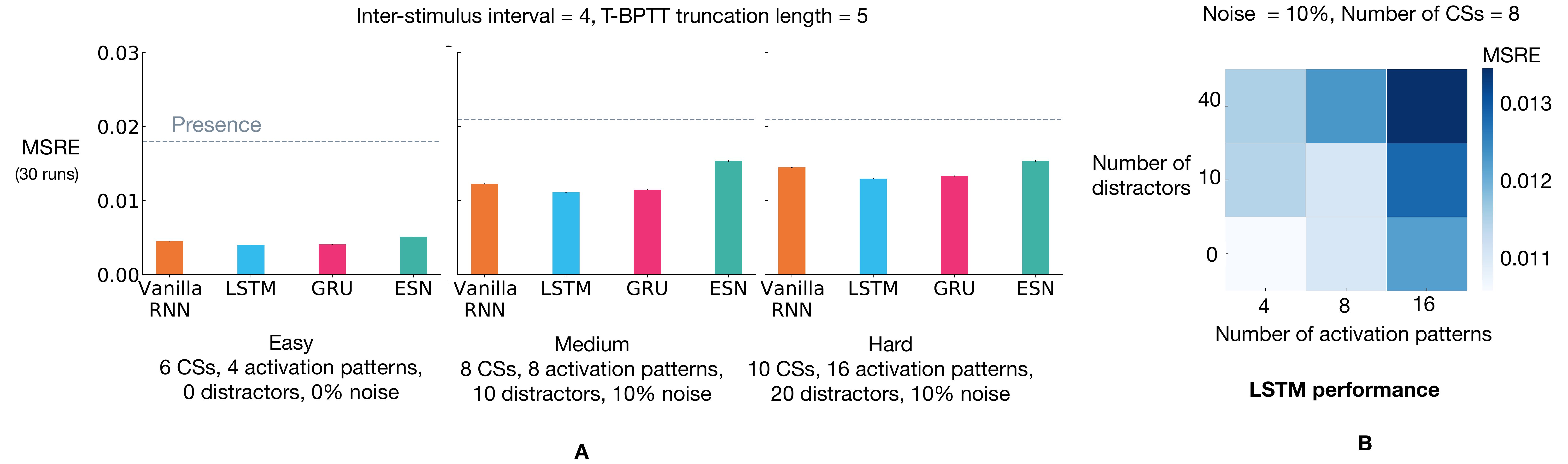}
  \caption{{\bf Noisy patterning} with varying difficulty levels.
The 3 bar plots (A) show the MSRE of Vanilla-RNN, GRU, and LSTM trained with T-BPTT as well as the MSRE of echo state network for three different configurations of the problem: easy, medium, and hard. 
The results are for 2 million steps of training and averaged over 30 runs.
The standard error bars are included. 
In the leftmost plot, we see a consistent drop in performance, across all methods, from the easy setting to the hard one. 
The heat map on the right (B), illustrates that the performance of LSTM degraded as the the number of distractors and activation patterns increased.
  }
   \label{fig:problem_2}
\end{figure*}

Just as we can control the difficulty level of trace conditioning by changing, for example, the ISI, we can also control the difficulty level of noisy patterning by changing the key problem parameters --- the number of CSs, the number of activation patterns, the number of distractors, and the level of noise.
Using this flexibility, we experimented with noisy patterning in two ways. 
First, we evaluated echo state network and several T-BPTT variants with truncation length $5$ on three different levels of difficulty that we refer to as easy, medium, and hard.

We did not experiment with RTRL because with small ISI ($=4$), T-BPTT with $T=5$ performs as well as the idealized RTRL baseline. 
We also did not experiment with tile-coded traces and microstimulus because they independently represent each input and cannot predict patterns of CSs as they are combined with linear function approximation.

There was a consistent drop in performance, across all methods, as the level of difficulty was increased (Figure \ref{fig:problem_2}, A). Echo state network performed worse than all three recurrent variants trained with T-BPTT in all three configurations of the problem. This is likely due to the fact that echo state network's representation, which is randomly determined and fixed at the beginning of learning, is not suitable for capturing the activation patterns.

\begin{figure}
\centering
  \includegraphics[width=0.5\linewidth]{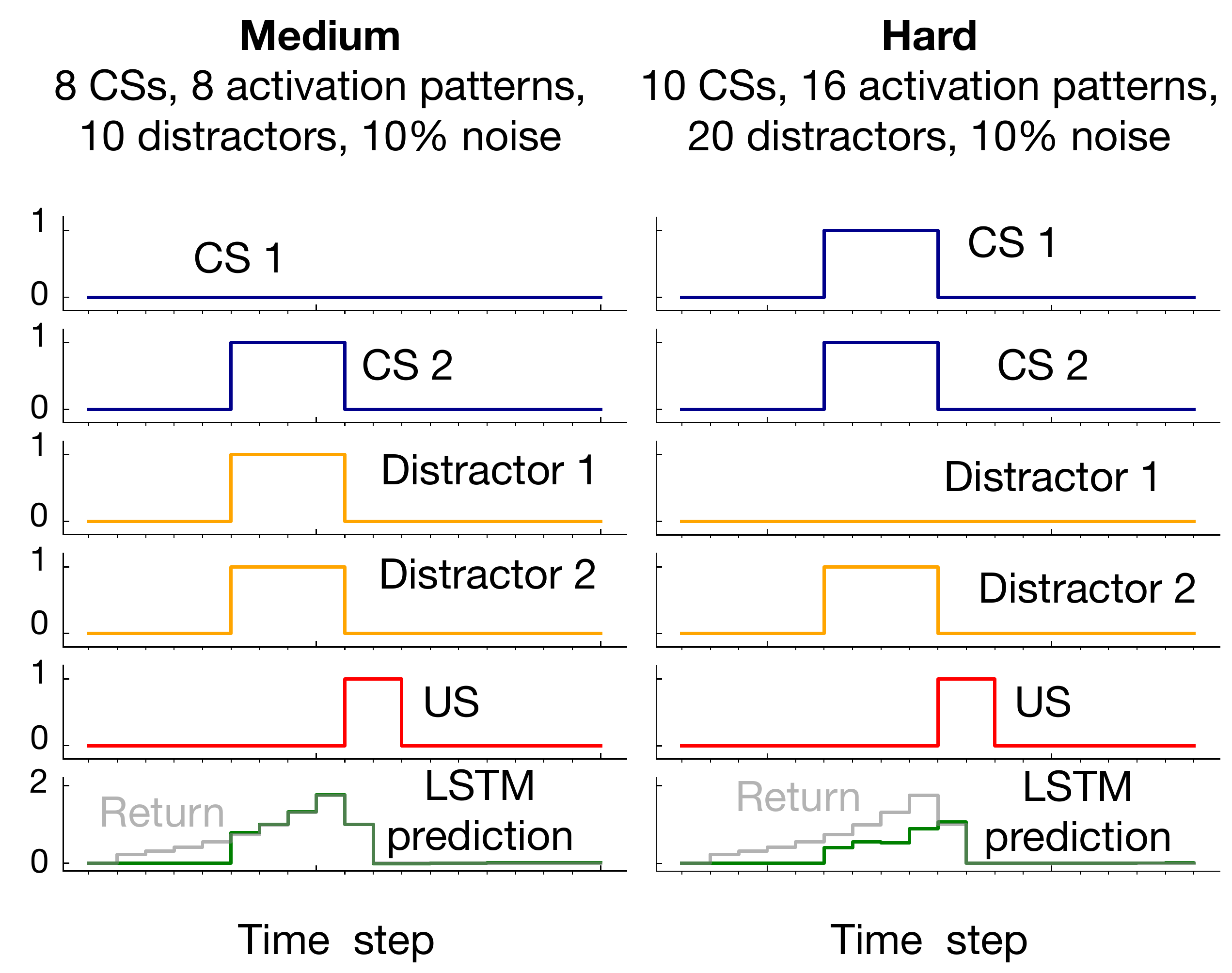}
  \caption{Example prediction profile plots for noisy patterning in the medium setting and hard setting. 
  Unlike Figure \ref{fig:problem_2_signals} where all the CSs and distractors were shown, in this figure only two of the CSs and distractors are shown as examples. In both cases, an activation pattern occurred as a result of which the US got activated. In the the medium setting, LSTM prediction matched the return. In the hard setting, however, LSTM did not predict the US accurately.}
   \label{fig:problem_2_predictions.pdf}
\end{figure}

Example prediction profile plots for noisy patterning are provided in Figure \ref{fig:problem_2_predictions.pdf} for the medium and hard levels of difficulty. We are only showing $2$ of the CSs and $2$ of the distractors as examples. In both examples, an activation pattern occurred and the US got activated (i.e., the US activation was not due to noise). In the medium setting, LSTM successfully predicted the US, matching the return after the onset of the CS. However, in the hard setting, there was a mismatch between LSTM's prediction and the return.

To further highlight the configurability of noisy patterning, we evaluated the T-BPTT variant of LSTM across two dimensions: the number of activation patterns and the number of distractors.
The results, presented as a heatmap of MSRE in Figure \ref{fig:problem_2} (B), show that the performance deteriorated as we made the problem more difficult across either dimension.

Taken together, these results demonstrate that noisy patterning can be useful for systematically studying the scaling properties of the algorithms in isolation from the temporal dimension, by simply increasing the number of signals from half a dozen to tens of thousands.

\begin{figure*}[]
\centering
  \includegraphics[width=0.8\linewidth]{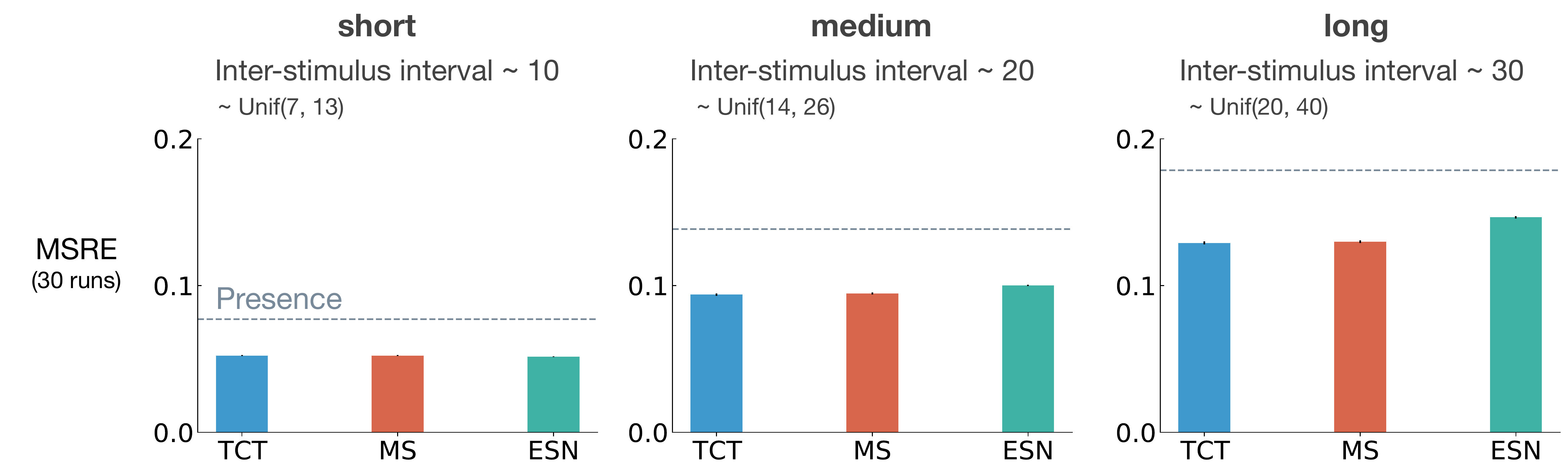}
  \caption{The impact of truncation level in {\bf trace patterning} for fixed representations. We used the exact same scheme as Figure \ref{fig:domain1-fixed} to visualize the performance in trace patterning. 
  Each plot corresponds to one setting of short, medium, and long ISI.
  Each bar reports the MSRE averaged over 30 runs.  
  All methods were trained for 5 million steps. 
  All fixed representations performed poorly.
  Tile-coded traces and microstimulus independently represent each input (not combinations) and thus cannot learn accurate predictions.
  }
   \label{fig:domain3-fixed}
\end{figure*}

\section{Trace Patterning: Putting It All Together}
\label{section:Test problem 3}
We put together the challenge of bridging the temporal gap, as posed by trace conditioning, and the challenge of recognizing important patterns, as formulated in noisy patterning, in a unified diagnostic problem that we refer to as \textit{trace patterning}.
For a learner to do well on this problem, it has to both fill the trace interval and construct non-linear representations of the CSs.

\begin{figure*}[]
\centering
  \includegraphics[width=0.8\linewidth]{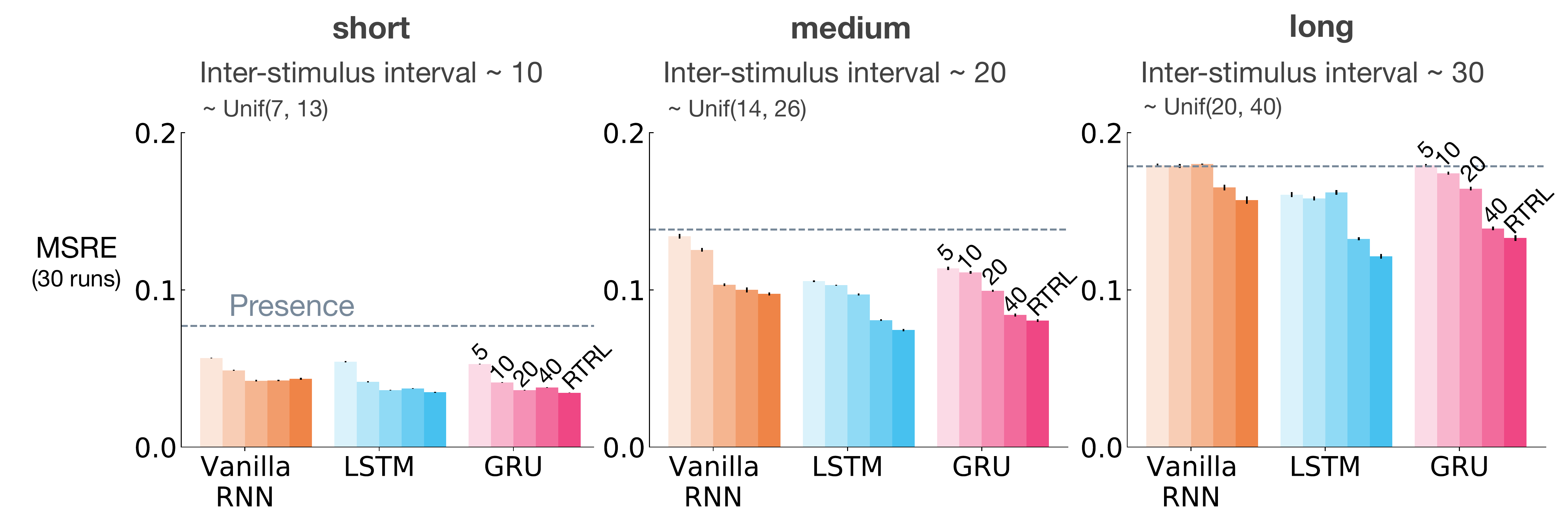}
  \caption{The impact of truncation level in {\bf trace patterning} for representations learned by T-BPTT and RTRL. 
  Each subplot corresponds to one setting of short, medium, and long ISI and includes the error for Vanilla-RNN, LSTM, and GRU.
  For each architecture, multiple bars are shown with the left four bars corresponding to T-BPTT with different $T$'s and the right bar corresponding to RTRL. 
  The results are calculated over 5 million steps and averaged over 30 runs. 
 Similar to trace conditioning, the T-BPTT based methods showed sensitivity to the truncation parameter. 
  The use of RTRL always improved performance; however, except for ISI$\sim$10 no methods learned accurate predictions. 
  }
   \label{fig:domain3-rnn}
\end{figure*}

Similar to the results presented in Section \ref{sec:test_problem_1}, we evaluated the baseline methods as we increased the ISI while keeping the rest of the problem parameters constant (8 CSs, 8 activation patterns, 10 distractors, and 10\% noise). 
The results for fixed representations and representations learned by T-BPTT and RTRL are provided in Figure \ref{fig:domain3-fixed} and \ref{fig:domain3-rnn} respectively.

The fixed representations performed poorly in all cases of short, medium, and long ISI and their performance got worse as ISI got larger (Figure \ref{fig:domain3-fixed}).  
The expert designed fixed representations of microstimulus and tile-coded traces independently represent each input (and not their combinations) and thus cannot learn accurate predictions; 
contextualizing the failure of the echo state network in this problem. 

The T-BPTT algorithms showed sensitivity to the length of the truncation window (Figure \ref{fig:domain3-rnn}). 
This is consistent with the findings from the trace conditioning experiments. 
One key difference, however, is that longer truncation parameter for the LSTM and GRU variants did not help as much as in trace conditioning. 
Moreover, in contrast to trace conditioning, the performance of the idealized RTRL baselines for the LSTM and GRU variants got worse considerably as we increased the ISI. 

\begin{figure*}
\centering
  \includegraphics[width=0.7\linewidth]{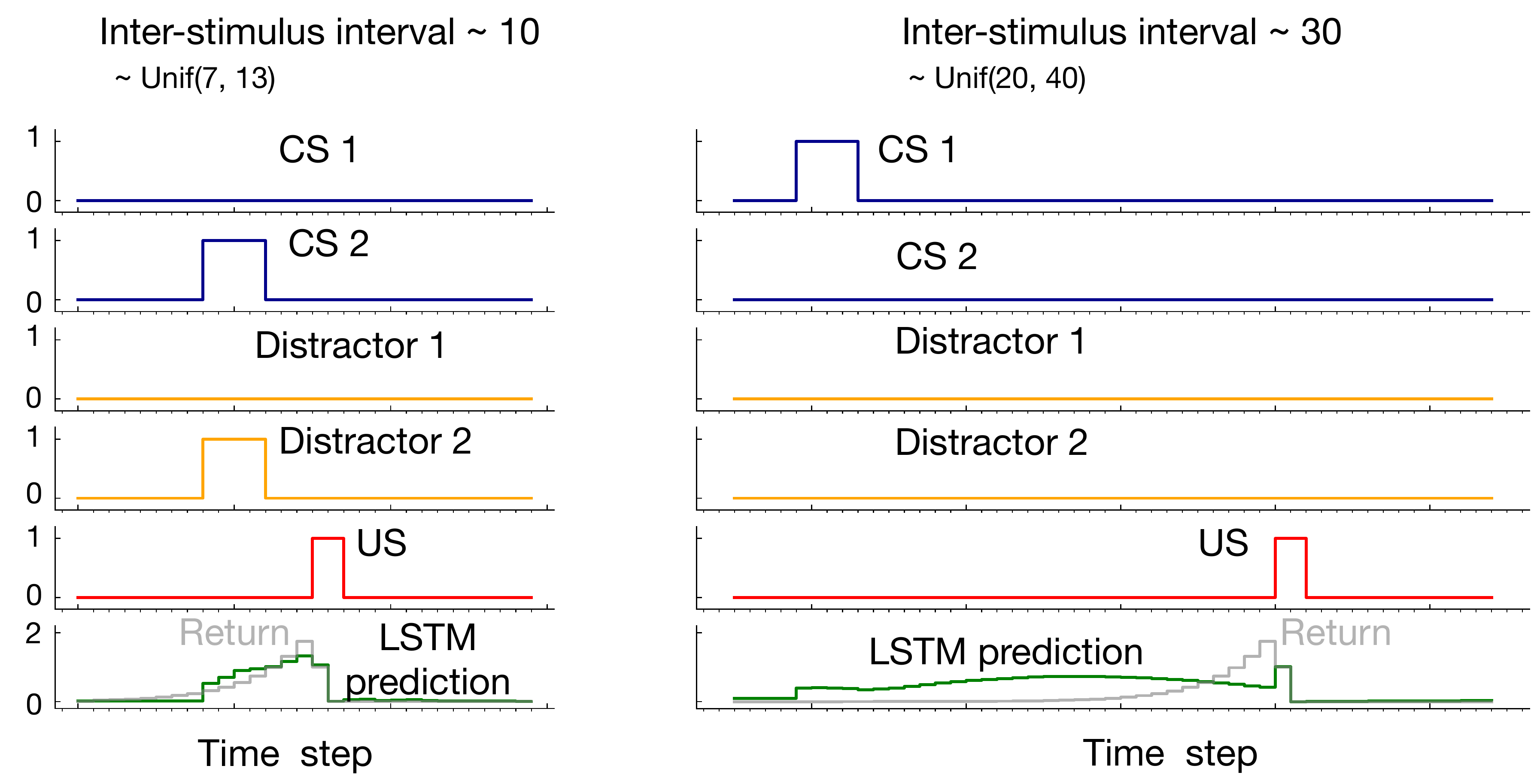}
  \caption{Example prediction profile plots for LSTM in {\bf trace patterning} in the the case of expected ISI $10$ and $30$. LSTM was trained with T-BPTT and a truncation length of $40$. Only two of the CS and distractors are shown as examples. In both cases, an activation pattern occurred as a result of which the US got activated. In the the case of expected ISI of $10$, LSTM prediction resembled the return. In the case of longer ISI with expectation of $30$, however, LSTM did not predict the US accurately.}
   \label{fig:problem_3_predictions}
\end{figure*}

Example prediction plots for LSTM trained with T-BPTT are shown in Figure \ref{fig:problem_3_predictions} in the case of expected ISI of 10 and 30. In both cases a truncation length of $40$ was used. While LSTM prediction profiles resemble the return in the case of expected ISI of $10$, they fail to match the return in the case of expected ISI of $30$.

This result emphasizes the difficulty of trace patterning --- the tested recurrent networks struggle to achieve low error, even when they have access to better gradient approximations, as in the case of training with RTRL.

\section{Combining Stimulating Traces with RNNs}
Our experimental results highlight the limitations of the current learning methods. 
While the linear trace-based methods successfully bridge the temporal gap in trace conditioning, their performance deteriorates when we introduce nonlinearities in trace patterning.
On the other hand, the recurrent learning algorithms can simultaneously bridge the temporal gap and handle nonlinearities, but they can be expensive in computational and memory requirements

\begin{figure*}[]
\centering
  \includegraphics[width=0.8\linewidth]{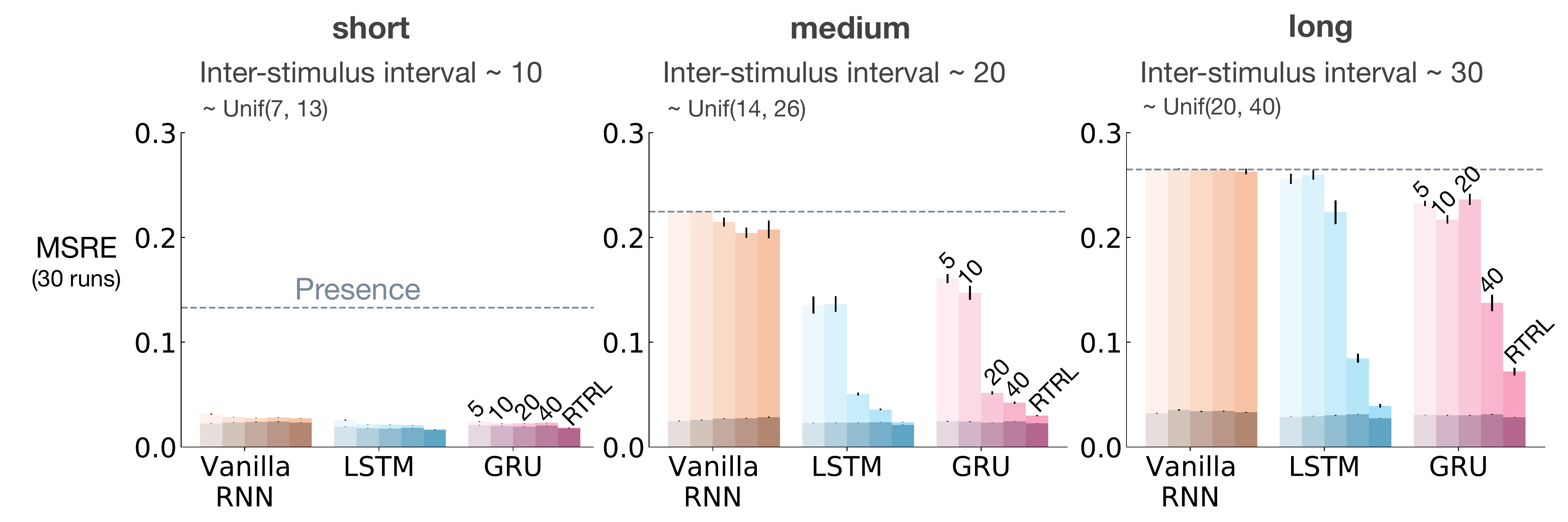}
  \caption{Results for combining stimulating traces with RNNs in {\bf trace conditioning}.
  We used the exact same scheme as Figure \ref{fig:domain1-rnn}.
  Darker colors denote the combination of stimulating traces with the recurrent methods and lighter shades denote the recurrent methods. 
  Each bar reports the MSRE averaged over 30 runs. 
  The methods were trained for 2 million steps.
  The error bars denote the standard errors. 
  Adding stimulating traces to the input of the Vanilla-RNN, GRU, and LSTM improved their performance in both T-BPTT and RTRL cases and made them less sensitive to the truncation length in the case of training with T-BPTT.
  }
   \label{fig:trace+RNN}
\end{figure*}

\begin{figure*}[]
\centering
  \includegraphics[width=0.8\linewidth]{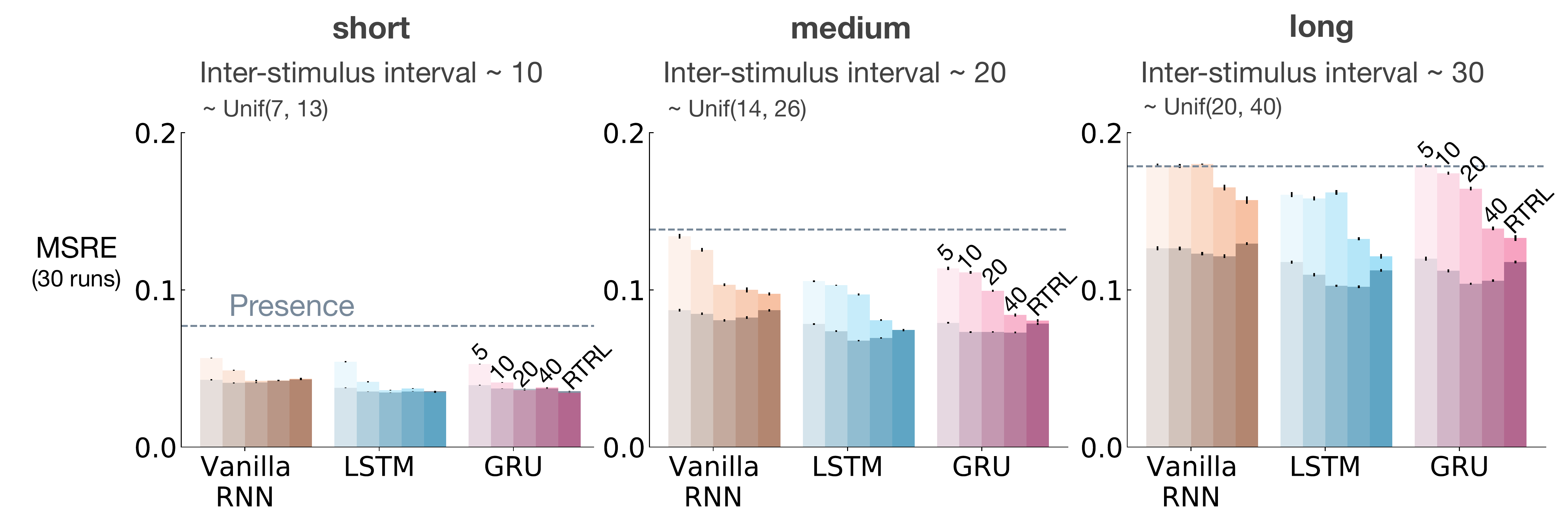}
  \caption{Results for combining stimulating traces with RNNs in {\bf trace patterning}. The naming conventions exactly match Figure \ref{fig:trace+RNN}, as does the general conclusion that stimulating traces improve performance but less so than in trace conditioning.
  }
   \label{fig:P3+RNN}
\end{figure*}

In the case of T-BPTT, the memory requirements of RNNs grow with the length of the truncation window, and learning long-term dependencies, as in trace conditioning, requires a comparably long truncation window. 
In the case of RTRL,  the computational complexity of RNNs grows quartically in the size of the hidden state, and learning patterns from a large number of signals, as in noisy patterning, requires a large hidden state.
Ideally, we need training methods that scale well in computation and memory simultaneously. 

As an example, we present a simple approach that scales well in computation and memory.
We augment the RNNs with the stimulating memory traces of the observation.
In particular, we feed an exponentially decaying trace of each stimulus, as described in Section \ref{sec:background}, as part of the input observation to the recurrent network. 

Figure \ref{fig:trace+RNN} and \ref{fig:P3+RNN} show the effect of augmenting the RNNs with the stimulating memory traces of the observation respectively in trace conditioning and trace patterning. 
The results for RNNs fed with only the observation is also included in lighter shades for comparison. 

When trained with T-BPTT, feeding the RNNs with the stimulating traces significantly improved the performance for the Vanilla RNN, LSTM, and GRU variants in trace conditioning. 
Moreover, it made the T-BPTT variants robust to the truncation length, achieving a similar level of error for all $T$'s. 
This effect was more pronounced in the long setting (Figure \ref{fig:trace+RNN}).  

When trained with RTRL, feeding the RNNs with the stimulating traces helped improve the performance (Figure \ref{fig:trace+RNN}). The improvement was larger for Vanilla RNN than the LSTM and GRU variants.  

In trace patterning, also feeding the RNNs with the stimulating traces improved performance in both T-BPTT and RTRL variants but less so than in trace conditioning. 

While the space of ideas for fruitfully combining memory traces and RNNs needs further investigation, this result shows how the proposed diagnostic benchmarks can help us search for general and scalable ideas for the online prediction problem. 

\section{Discussion}
Our diagnostic benchmarks can be used to isolate and investigate fundamental algorithmic issues in recurrent learning.
In trace conditioning, we found that basic recurrent architectures could not handle significant temporal dependencies. Gated architectures exhibited significant sensitivity to truncation level (needing to unroll beyond the onset of the relevant cue) but did not perform as well as RTRL variants. 
In our trace patterning experiments, all methods struggled when confronted with the combination of long temporal dependencies and the need to extract configuration patterns.

In this paper, we investigated the online prediction setting, but more stringent computational restrictions might be useful for future work.
Many RL algorithms, like TD, can make and update long-horizon predictions with computation significantly less than the length of the prediction's horizon \citep{van2015learning}.
This might also be possible in representation learning. 
Can the agent construct representations capable of overcoming dependencies back in time with computation and storage less than the length of the gap?
While recurrent learning algorithms based solely on T-BPTT do not meet this requirement, our results show that some combination of stimulating traces and recurrent architectures may reduce the agent's dependency on the truncation level.
Moreover, there is a discrepancy between the speed of learning for natural and artificial systems; 
while animals learn eyeblink conditioning in about a few hundred trials, our baseline methods require thousands of trials to learn the task.
Future research should investigate reasonable computational restrictions if we hope to discover representations as efficient as those used by animals. Work on more efficient update rules \citep{nath2019training} and attention mechanisms \citep{dehghani2018universal} represent promising directions toward this ambitious goal.

\section{Acknowledgement}
We would like to thank Martha White for many helpful discussions. We also thank Patrick Pilarski, Doina Precup, and Khurram Javed for providing helpful comments for improving the paper.


\bibliography{paper}

\begin{thebibliography}{}

\bibitem [\protect \citeauthoryear {%
Baird%
}{%
Baird%
}{%
{\protect \APACyear {1995}}%
}]{%
baird1995residual}
\APACinsertmetastar {%
baird1995residual}%
\begin{APACrefauthors}%
Baird, L.%
\end{APACrefauthors}%
\unskip\
\newblock
\APACrefYearMonthDay{1995}{}{}.
\newblock
{\BBOQ}\APACrefatitle {Residual algorithms: Reinforcement learning with
  function approximation} {Residual algorithms: Reinforcement learning with
  function approximation}.{\BBCQ}
\newblock
\BIn{} \APACrefbtitle {Machine Learning Proceedings 1995} {Machine learning
  proceedings 1995}\ (\BPGS\ 30--37).
\newblock
\APACaddressPublisher{}{Elsevier}.
\PrintBackRefs{\CurrentBib}

\bibitem [\protect \citeauthoryear {%
Beattie%
\ \protect \BOthers {.}}{%
Beattie%
\ \protect \BOthers {.}}{%
{\protect \APACyear {2016}}%
}]{%
beattie2016deepmind}
\APACinsertmetastar {%
beattie2016deepmind}%
\begin{APACrefauthors}%
Beattie, C.%
, Leibo, J\BPBI Z.%
, Teplyashin, D.%
, Ward, T.%
, Wainwright, M.%
, K{\"u}ttler, H.%
\BDBL {}Petersen, S.%
\end{APACrefauthors}%
\unskip\
\newblock
\APACrefYearMonthDay{2016}{}{}.
\newblock
{\BBOQ}\APACrefatitle {Deepmind lab} {Deepmind lab}.{\BBCQ}
\newblock
\APACjournalVolNumPages{arXiv preprint arXiv:1612.03801}{}{}{}.
\PrintBackRefs{\CurrentBib}

\bibitem [\protect \citeauthoryear {%
Bellemare%
, Naddaf%
, Veness%
\BCBL {}\ \BBA {} Bowling%
}{%
Bellemare%
\ \protect \BOthers {.}}{%
{\protect \APACyear {2013}}%
}]{%
bellemare2013arcade}
\APACinsertmetastar {%
bellemare2013arcade}%
\begin{APACrefauthors}%
Bellemare, M\BPBI G.%
, Naddaf, Y.%
, Veness, J.%
\BCBL {}\ \BBA {} Bowling, M.%
\end{APACrefauthors}%
\unskip\
\newblock
\APACrefYearMonthDay{2013}{}{}.
\newblock
{\BBOQ}\APACrefatitle {The arcade learning environment: An evaluation platform
  for general agents} {The arcade learning environment: An evaluation platform
  for general agents}.{\BBCQ}
\newblock
\APACjournalVolNumPages{Journal of Artificial Intelligence
  Research}{47}{}{253--279}.
\PrintBackRefs{\CurrentBib}

\bibitem [\protect \citeauthoryear {%
Brockman%
\ \protect \BOthers {.}}{%
Brockman%
\ \protect \BOthers {.}}{%
{\protect \APACyear {2016}}%
}]{%
brockman2016openai}
\APACinsertmetastar {%
brockman2016openai}%
\begin{APACrefauthors}%
Brockman, G.%
, Cheung, V.%
, Pettersson, L.%
, Schneider, J.%
, Schulman, J.%
, Tang, J.%
\BCBL {}\ \BBA {} Zaremba, W.%
\end{APACrefauthors}%
\unskip\
\newblock
\APACrefYearMonthDay{2016}{}{}.
\newblock
{\BBOQ}\APACrefatitle {Openai gym} {Openai gym}.{\BBCQ}
\newblock
\APACjournalVolNumPages{arXiv preprint arXiv:1606.01540}{}{}{}.
\PrintBackRefs{\CurrentBib}

\bibitem [\protect \citeauthoryear {%
Ceron%
\ \BBA {} Castro%
}{%
Ceron%
\ \BBA {} Castro%
}{%
{\protect \APACyear {2021}}%
}]{%
ceron2021revisiting}
\APACinsertmetastar {%
ceron2021revisiting}%
\begin{APACrefauthors}%
Ceron, J\BPBI S\BPBI O.%
\BCBT {}\ \BBA {} Castro, P\BPBI S.%
\end{APACrefauthors}%
\unskip\
\newblock
\APACrefYearMonthDay{2021}{}{}.
\newblock
{\BBOQ}\APACrefatitle {Revisiting rainbow: Promoting more insightful and
  inclusive deep reinforcement learning research} {Revisiting rainbow:
  Promoting more insightful and inclusive deep reinforcement learning
  research}.{\BBCQ}
\newblock
\BIn{} \APACrefbtitle {International Conference on Machine Learning}
  {International conference on machine learning}\ (\BPGS\ 1373--1383).
\PrintBackRefs{\CurrentBib}

\bibitem [\protect \citeauthoryear {%
Chen%
\ \protect \BOthers {.}}{%
Chen%
\ \protect \BOthers {.}}{%
{\protect \APACyear {2021}}%
}]{%
chen2021decision}
\APACinsertmetastar {%
chen2021decision}%
\begin{APACrefauthors}%
Chen, L.%
, Lu, K.%
, Rajeswaran, A.%
, Lee, K.%
, Grover, A.%
, Laskin, M.%
\BDBL {}Mordatch, I.%
\end{APACrefauthors}%
\unskip\
\newblock
\APACrefYearMonthDay{2021}{}{}.
\newblock
{\BBOQ}\APACrefatitle {Decision transformer: Reinforcement learning via
  sequence modeling} {Decision transformer: Reinforcement learning via sequence
  modeling}.{\BBCQ}
\newblock
\APACjournalVolNumPages{Advances in neural information processing
  systems}{34}{}{}.
\PrintBackRefs{\CurrentBib}

\bibitem [\protect \citeauthoryear {%
Cho%
, Van~Merri{\"e}nboer%
, Bahdanau%
\BCBL {}\ \BBA {} Bengio%
}{%
Cho%
\ \protect \BOthers {.}}{%
{\protect \APACyear {2014}}%
}]{%
cho2014properties}
\APACinsertmetastar {%
cho2014properties}%
\begin{APACrefauthors}%
Cho, K.%
, Van~Merri{\"e}nboer, B.%
, Bahdanau, D.%
\BCBL {}\ \BBA {} Bengio, Y.%
\end{APACrefauthors}%
\unskip\
\newblock
\APACrefYearMonthDay{2014}{}{}.
\newblock
{\BBOQ}\APACrefatitle {On the properties of neural machine translation:
  Encoder-decoder approaches} {On the properties of neural machine translation:
  Encoder-decoder approaches}.{\BBCQ}
\newblock
\APACjournalVolNumPages{arXiv preprint arXiv:1409.1259}{}{}{}.
\PrintBackRefs{\CurrentBib}

\bibitem [\protect \citeauthoryear {%
Colas%
, Sigaud%
\BCBL {}\ \BBA {} Oudeyer%
}{%
Colas%
\ \protect \BOthers {.}}{%
{\protect \APACyear {2018}}%
}]{%
colas2018many}
\APACinsertmetastar {%
colas2018many}%
\begin{APACrefauthors}%
Colas, C.%
, Sigaud, O.%
\BCBL {}\ \BBA {} Oudeyer, P.%
\end{APACrefauthors}%
\unskip\
\newblock
\APACrefYearMonthDay{2018}{}{}.
\newblock
{\BBOQ}\APACrefatitle {How many random seeds? Statistical power analysis in
  deep reinforcement learning experiments} {How many random seeds? statistical
  power analysis in deep reinforcement learning experiments}.{\BBCQ}
\newblock
\APACjournalVolNumPages{arXiv preprint arXiv:1806.08295}{}{}{}.
\PrintBackRefs{\CurrentBib}

\bibitem [\protect \citeauthoryear {%
Dehghani%
, Gouws%
, Vinyals%
, Uszkoreit%
\BCBL {}\ \BBA {} Kaiser%
}{%
Dehghani%
\ \protect \BOthers {.}}{%
{\protect \APACyear {2019}}%
}]{%
dehghani2018universal}
\APACinsertmetastar {%
dehghani2018universal}%
\begin{APACrefauthors}%
Dehghani, M.%
, Gouws, S.%
, Vinyals, O.%
, Uszkoreit, J.%
\BCBL {}\ \BBA {} Kaiser, L.%
\end{APACrefauthors}%
\unskip\
\newblock
\APACrefYearMonthDay{2019}{}{}.
\newblock
{\BBOQ}\APACrefatitle {Universal Transformers} {Universal transformers}.{\BBCQ}
\newblock
\BIn{} \APACrefbtitle {International Conference on Learning Representations.}
  {International conference on learning representations.}
\PrintBackRefs{\CurrentBib}

\bibitem [\protect \citeauthoryear {%
Dickinson%
}{%
Dickinson%
}{%
{\protect \APACyear {1980}}%
}]{%
dickinson1980contemporary}
\APACinsertmetastar {%
dickinson1980contemporary}%
\begin{APACrefauthors}%
Dickinson, A.%
\end{APACrefauthors}%
\unskip\
\newblock
\APACrefYear{1980}.
\newblock
\APACrefbtitle {Contemporary animal learning theory} {Contemporary animal
  learning theory}\ (\BVOL~1).
\newblock
\APACaddressPublisher{}{CUP Archive}.
\PrintBackRefs{\CurrentBib}

\bibitem [\protect \citeauthoryear {%
Elman%
}{%
Elman%
}{%
{\protect \APACyear {1990}}%
}]{%
elman1990finding}
\APACinsertmetastar {%
elman1990finding}%
\begin{APACrefauthors}%
Elman, J\BPBI L.%
\end{APACrefauthors}%
\unskip\
\newblock
\APACrefYearMonthDay{1990}{}{}.
\newblock
{\BBOQ}\APACrefatitle {Finding structure in time} {Finding structure in
  time}.{\BBCQ}
\newblock
\APACjournalVolNumPages{Cognitive science}{14}{2}{179--211}.
\PrintBackRefs{\CurrentBib}

\bibitem [\protect \citeauthoryear {%
Engstrom%
\ \protect \BOthers {.}}{%
Engstrom%
\ \protect \BOthers {.}}{%
{\protect \APACyear {2019}}%
}]{%
engstrom2019implementation}
\APACinsertmetastar {%
engstrom2019implementation}%
\begin{APACrefauthors}%
Engstrom, L.%
, Ilyas, A.%
, Santurkar, S.%
, Tsipras, D.%
, Janoos, F.%
, Rudolph, L.%
\BCBL {}\ \BBA {} Madry, A.%
\end{APACrefauthors}%
\unskip\
\newblock
\APACrefYearMonthDay{2019}{}{}.
\newblock
{\BBOQ}\APACrefatitle {Implementation matters in deep rl: A case study on ppo
  and trpo} {Implementation matters in deep rl: A case study on ppo and
  trpo}.{\BBCQ}
\newblock
\BIn{} \APACrefbtitle {International Conference on Learning Representations.}
  {International conference on learning representations.}
\PrintBackRefs{\CurrentBib}

\bibitem [\protect \citeauthoryear {%
Espeholt%
\ \protect \BOthers {.}}{%
Espeholt%
\ \protect \BOthers {.}}{%
{\protect \APACyear {2018}}%
}]{%
espeholt2018impala}
\APACinsertmetastar {%
espeholt2018impala}%
\begin{APACrefauthors}%
Espeholt, L.%
, Soyer, H.%
, Munos, R.%
, Simonyan, K.%
, Mnih, V.%
, Ward, T.%
\BDBL {}Kavukcuoglu, K.%
\end{APACrefauthors}%
\unskip\
\newblock
\APACrefYearMonthDay{2018}{10--15 Jul}{}.
\newblock
{\BBOQ}\APACrefatitle {{IMPALA}: Scalable Distributed Deep-{RL} with Importance
  Weighted Actor-Learner Architectures} {{IMPALA}: Scalable distributed
  deep-{RL} with importance weighted actor-learner architectures}.{\BBCQ}
\newblock
\BIn{} \APACrefbtitle {Proceedings of the 35th International Conference on
  Machine Learning} {Proceedings of the 35th international conference on
  machine learning}\ (\BVOL~80, \BPGS\ 1407--1416).
\newblock
\APACaddressPublisher{Stockholmsmässan, Stockholm Sweden}{PMLR}.
\PrintBackRefs{\CurrentBib}

\bibitem [\protect \citeauthoryear {%
Fortunato%
\ \protect \BOthers {.}}{%
Fortunato%
\ \protect \BOthers {.}}{%
{\protect \APACyear {2019}}%
}]{%
fortunato2019generalization}
\APACinsertmetastar {%
fortunato2019generalization}%
\begin{APACrefauthors}%
Fortunato, M.%
, Tan, M.%
, Faulkner, R.%
, Hansen, S.%
, Badia, A\BPBI P.%
, Buttimore, G.%
\BDBL {}Blundell, C.%
\end{APACrefauthors}%
\unskip\
\newblock
\APACrefYearMonthDay{2019}{}{}.
\newblock
{\BBOQ}\APACrefatitle {Generalization of reinforcement learners with working
  and episodic memory} {Generalization of reinforcement learners with working
  and episodic memory}.{\BBCQ}
\newblock
\BIn{} \APACrefbtitle {Advances in Neural Information Processing Systems}
  {Advances in neural information processing systems}\ (\BPGS\ 12469--12478).
\PrintBackRefs{\CurrentBib}

\bibitem [\protect \citeauthoryear {%
Gallistel%
\ \BBA {} King%
}{%
Gallistel%
\ \BBA {} King%
}{%
{\protect \APACyear {2011}}%
}]{%
gallistel2011memory}
\APACinsertmetastar {%
gallistel2011memory}%
\begin{APACrefauthors}%
Gallistel, C\BPBI R.%
\BCBT {}\ \BBA {} King, A\BPBI P.%
\end{APACrefauthors}%
\unskip\
\newblock
\APACrefYear{2011}.
\newblock
\APACrefbtitle {Memory and the computational brain: Why cognitive science will
  transform neuroscience} {Memory and the computational brain: Why cognitive
  science will transform neuroscience}\ (\BVOL~6).
\newblock
\APACaddressPublisher{}{John Wiley \& Sons}.
\PrintBackRefs{\CurrentBib}

\bibitem [\protect \citeauthoryear {%
Gehring%
, Auli%
, Grangier%
, Yarats%
\BCBL {}\ \BBA {} Dauphin%
}{%
Gehring%
\ \protect \BOthers {.}}{%
{\protect \APACyear {2017}}%
}]{%
gehring2017convolutional}
\APACinsertmetastar {%
gehring2017convolutional}%
\begin{APACrefauthors}%
Gehring, J.%
, Auli, M.%
, Grangier, D.%
, Yarats, D.%
\BCBL {}\ \BBA {} Dauphin, Y\BPBI N.%
\end{APACrefauthors}%
\unskip\
\newblock
\APACrefYearMonthDay{2017}{}{}.
\newblock
{\BBOQ}\APACrefatitle {Convolutional sequence to sequence learning}
  {Convolutional sequence to sequence learning}.{\BBCQ}
\newblock
\BIn{} \APACrefbtitle {International Conference on Machine Learning}
  {International conference on machine learning}\ (\BPGS\ 1243--1252).
\PrintBackRefs{\CurrentBib}

\bibitem [\protect \citeauthoryear {%
Harris%
, Livesey%
, Gharaei%
\BCBL {}\ \BBA {} Westbrook%
}{%
Harris%
\ \protect \BOthers {.}}{%
{\protect \APACyear {2008}}%
}]{%
harris2008negative}
\APACinsertmetastar {%
harris2008negative}%
\begin{APACrefauthors}%
Harris, J\BPBI A.%
, Livesey, E\BPBI J.%
, Gharaei, S.%
\BCBL {}\ \BBA {} Westbrook, R\BPBI F.%
\end{APACrefauthors}%
\unskip\
\newblock
\APACrefYearMonthDay{2008}{}{}.
\newblock
{\BBOQ}\APACrefatitle {Negative patterning is easier than a biconditional
  discrimination.} {Negative patterning is easier than a biconditional
  discrimination.}{\BBCQ}
\newblock
\APACjournalVolNumPages{Journal of Experimental Psychology: Animal Behavior
  Processes}{34}{4}{494}.
\PrintBackRefs{\CurrentBib}

\bibitem [\protect \citeauthoryear {%
Henderson%
\ \protect \BOthers {.}}{%
Henderson%
\ \protect \BOthers {.}}{%
{\protect \APACyear {2018}}%
}]{%
henderson2018deep}
\APACinsertmetastar {%
henderson2018deep}%
\begin{APACrefauthors}%
Henderson, P.%
, Islam, R.%
, Bachman, P.%
, Pineau, J.%
, Precup, D.%
\BCBL {}\ \BBA {} Meger, D.%
\end{APACrefauthors}%
\unskip\
\newblock
\APACrefYearMonthDay{2018}{}{}.
\newblock
{\BBOQ}\APACrefatitle {Deep reinforcement learning that matters} {Deep
  reinforcement learning that matters}.{\BBCQ}
\newblock
\BIn{} \APACrefbtitle {Proceedings of the Association for the Advancement of
  Artificial Intelligence} {Proceedings of the association for the advancement
  of artificial intelligence}\ (\BVOL~32).
\PrintBackRefs{\CurrentBib}

\bibitem [\protect \citeauthoryear {%
Hochreiter%
, Bengio%
, Frasconi%
\BCBL {}\ \BBA {} Schmidhuber%
}{%
Hochreiter%
\ \protect \BOthers {.}}{%
{\protect \APACyear {2001}}%
}]{%
hochreiter2001gradient}
\APACinsertmetastar {%
hochreiter2001gradient}%
\begin{APACrefauthors}%
Hochreiter, S.%
, Bengio, Y.%
, Frasconi, P.%
\BCBL {}\ \BBA {} Schmidhuber, J.%
\end{APACrefauthors}%
\unskip\
\newblock
\APACrefYearMonthDay{2001}{}{}.
\newblock
{\BBOQ}\APACrefatitle {Gradient flow in recurrent nets: the difficulty of
  learning long-term dependencies} {Gradient flow in recurrent nets: the
  difficulty of learning long-term dependencies}.{\BBCQ}
\newblock
\APACjournalVolNumPages{A Field Guide to Dynamical Recurrent Neural
  Networks}{}{}{}.
\PrintBackRefs{\CurrentBib}

\bibitem [\protect \citeauthoryear {%
Hochreiter%
\ \BBA {} Schmidhuber%
}{%
Hochreiter%
\ \BBA {} Schmidhuber%
}{%
{\protect \APACyear {1997}}%
}]{%
hochreiter1997long}
\APACinsertmetastar {%
hochreiter1997long}%
\begin{APACrefauthors}%
Hochreiter, S.%
\BCBT {}\ \BBA {} Schmidhuber, J.%
\end{APACrefauthors}%
\unskip\
\newblock
\APACrefYearMonthDay{1997}{}{}.
\newblock
{\BBOQ}\APACrefatitle {Long short-term memory} {Long short-term memory}.{\BBCQ}
\newblock
\APACjournalVolNumPages{Neural Computation}{9}{8}{1735--1780}.
\PrintBackRefs{\CurrentBib}

\bibitem [\protect \citeauthoryear {%
Hopfield%
}{%
Hopfield%
}{%
{\protect \APACyear {1982}}%
}]{%
hopfield1982neural}
\APACinsertmetastar {%
hopfield1982neural}%
\begin{APACrefauthors}%
Hopfield, J\BPBI J.%
\end{APACrefauthors}%
\unskip\
\newblock
\APACrefYearMonthDay{1982}{}{}.
\newblock
{\BBOQ}\APACrefatitle {Neural networks and physical systems with emergent
  collective computational abilities} {Neural networks and physical systems
  with emergent collective computational abilities}.{\BBCQ}
\newblock
\APACjournalVolNumPages{Proceedings of the National Academy of
  Sciences}{79}{8}{2554--2558}.
\PrintBackRefs{\CurrentBib}

\bibitem [\protect \citeauthoryear {%
Howard%
\ \BBA {} Eichenbaum%
}{%
Howard%
\ \BBA {} Eichenbaum%
}{%
{\protect \APACyear {2013}}%
}]{%
howard2013hippocampus}
\APACinsertmetastar {%
howard2013hippocampus}%
\begin{APACrefauthors}%
Howard, M\BPBI W.%
\BCBT {}\ \BBA {} Eichenbaum, H.%
\end{APACrefauthors}%
\unskip\
\newblock
\APACrefYearMonthDay{2013}{}{}.
\newblock
{\BBOQ}\APACrefatitle {The hippocampus, time, and memory across scales.} {The
  hippocampus, time, and memory across scales.}{\BBCQ}
\newblock
\APACjournalVolNumPages{Journal of Experimental Psychology:
  General}{142}{4}{1211}.
\PrintBackRefs{\CurrentBib}

\bibitem [\protect \citeauthoryear {%
Hull%
}{%
Hull%
}{%
{\protect \APACyear {1939}}%
}]{%
hull1939problem}
\APACinsertmetastar {%
hull1939problem}%
\begin{APACrefauthors}%
Hull, C\BPBI L.%
\end{APACrefauthors}%
\unskip\
\newblock
\APACrefYearMonthDay{1939}{}{}.
\newblock
{\BBOQ}\APACrefatitle {The problem of stimulus equivalence in behavior theory.}
  {The problem of stimulus equivalence in behavior theory.}{\BBCQ}
\newblock
\APACjournalVolNumPages{Psychological Review}{46}{1}{9}.
\PrintBackRefs{\CurrentBib}

\bibitem [\protect \citeauthoryear {%
Jacobsen%
\ \protect \BOthers {.}}{%
Jacobsen%
\ \protect \BOthers {.}}{%
{\protect \APACyear {2019}}%
}]{%
jacobsen2019meta}
\APACinsertmetastar {%
jacobsen2019meta}%
\begin{APACrefauthors}%
Jacobsen, A.%
, Schlegel, M.%
, Linke, C.%
, Degris, T.%
, White, A.%
\BCBL {}\ \BBA {} White, M.%
\end{APACrefauthors}%
\unskip\
\newblock
\APACrefYearMonthDay{2019}{}{}.
\newblock
{\BBOQ}\APACrefatitle {Meta-descent for online, continual prediction}
  {Meta-descent for online, continual prediction}.{\BBCQ}
\newblock
\BIn{} \APACrefbtitle {Proceedings of the Association for the Advancement of
  Artificial Intelligence} {Proceedings of the association for the advancement
  of artificial intelligence}\ (\BVOL~33, \BPGS\ 3943--3950).
\PrintBackRefs{\CurrentBib}

\bibitem [\protect \citeauthoryear {%
Jaderberg%
\ \protect \BOthers {.}}{%
Jaderberg%
\ \protect \BOthers {.}}{%
{\protect \APACyear {2017}}%
}]{%
jaderberg2017decoupled}
\APACinsertmetastar {%
jaderberg2017decoupled}%
\begin{APACrefauthors}%
Jaderberg, M.%
, Czarnecki, W\BPBI M.%
, Osindero, S.%
, Vinyals, O.%
, Graves, A.%
, Silver, D.%
\BCBL {}\ \BBA {} Kavukcuoglu, K.%
\end{APACrefauthors}%
\unskip\
\newblock
\APACrefYearMonthDay{2017}{}{}.
\newblock
{\BBOQ}\APACrefatitle {Decoupled neural interfaces using synthetic gradients}
  {Decoupled neural interfaces using synthetic gradients}.{\BBCQ}
\newblock
\BIn{} \APACrefbtitle {International Conference on Machine Learning}
  {International conference on machine learning}\ (\BPGS\ 1627--1635).
\PrintBackRefs{\CurrentBib}

\bibitem [\protect \citeauthoryear {%
Jaeger%
}{%
Jaeger%
}{%
{\protect \APACyear {2001}}%
}]{%
jaeger2001echo}
\APACinsertmetastar {%
jaeger2001echo}%
\begin{APACrefauthors}%
Jaeger, H.%
\end{APACrefauthors}%
\unskip\
\newblock
\APACrefYearMonthDay{2001}{}{}.
\newblock
{\BBOQ}\APACrefatitle {The “echo state” approach to analysing and training
  recurrent neural networks-with an erratum note} {The “echo state”
  approach to analysing and training recurrent neural networks-with an erratum
  note}.{\BBCQ}
\newblock
\APACjournalVolNumPages{German National Research Center for Information
  Technology GMD Technical Report}{148}{34}{13}.
\PrintBackRefs{\CurrentBib}

\bibitem [\protect \citeauthoryear {%
James%
}{%
James%
}{%
{\protect \APACyear {1890}}%
}]{%
james2007principles}
\APACinsertmetastar {%
james2007principles}%
\begin{APACrefauthors}%
James, W.%
\end{APACrefauthors}%
\unskip\
\newblock
\APACrefYear{1890}.
\newblock
\APACrefbtitle {The Principles of Psychology} {The principles of psychology}\
  (\BVOL~1).
\newblock
\APACaddressPublisher{}{Henry Holt and Company}.
\PrintBackRefs{\CurrentBib}

\bibitem [\protect \citeauthoryear {%
Janner%
, Li%
\BCBL {}\ \BBA {} Levine%
}{%
Janner%
\ \protect \BOthers {.}}{%
{\protect \APACyear {2021}}%
}]{%
janner2021reinforcement}
\APACinsertmetastar {%
janner2021reinforcement}%
\begin{APACrefauthors}%
Janner, M.%
, Li, Q.%
\BCBL {}\ \BBA {} Levine, S.%
\end{APACrefauthors}%
\unskip\
\newblock
\APACrefYearMonthDay{2021}{}{}.
\newblock
{\BBOQ}\APACrefatitle {Offline Reinforcement Learning as One Big Sequence
  Modeling Problem} {Offline reinforcement learning as one big sequence
  modeling problem}.{\BBCQ}
\newblock
\APACjournalVolNumPages{Advances in neural information processing
  systems}{34}{}{}.
\PrintBackRefs{\CurrentBib}

\bibitem [\protect \citeauthoryear {%
Kingma%
\ \BBA {} Ba%
}{%
Kingma%
\ \BBA {} Ba%
}{%
{\protect \APACyear {2014}}%
}]{%
kingma2014adam}
\APACinsertmetastar {%
kingma2014adam}%
\begin{APACrefauthors}%
Kingma, D\BPBI P.%
\BCBT {}\ \BBA {} Ba, J.%
\end{APACrefauthors}%
\unskip\
\newblock
\APACrefYearMonthDay{2014}{}{}.
\newblock
{\BBOQ}\APACrefatitle {Adam: A method for stochastic optimization} {Adam: A
  method for stochastic optimization}.{\BBCQ}
\newblock
\APACjournalVolNumPages{arXiv preprint arXiv:1412.6980}{}{}{}.
\PrintBackRefs{\CurrentBib}

\bibitem [\protect \citeauthoryear {%
Loynd%
, Fernandez%
, Celikyilmaz%
, Swaminathan%
\BCBL {}\ \BBA {} Hausknecht%
}{%
Loynd%
\ \protect \BOthers {.}}{%
{\protect \APACyear {2020}}%
}]{%
loynd2020working}
\APACinsertmetastar {%
loynd2020working}%
\begin{APACrefauthors}%
Loynd, R.%
, Fernandez, R.%
, Celikyilmaz, A.%
, Swaminathan, A.%
\BCBL {}\ \BBA {} Hausknecht, M.%
\end{APACrefauthors}%
\unskip\
\newblock
\APACrefYearMonthDay{2020}{}{}.
\newblock
{\BBOQ}\APACrefatitle {Working memory graphs} {Working memory graphs}.{\BBCQ}
\newblock
\BIn{} \APACrefbtitle {International Conference on Machine Learning}
  {International conference on machine learning}\ (\BPGS\ 6404--6414).
\PrintBackRefs{\CurrentBib}

\bibitem [\protect \citeauthoryear {%
Ludvig%
, Sutton%
\BCBL {}\ \BBA {} Kehoe%
}{%
Ludvig%
\ \protect \BOthers {.}}{%
{\protect \APACyear {2008}}%
}]{%
ludvig2008stimulus}
\APACinsertmetastar {%
ludvig2008stimulus}%
\begin{APACrefauthors}%
Ludvig, E\BPBI A.%
, Sutton, R\BPBI S.%
\BCBL {}\ \BBA {} Kehoe, E\BPBI J.%
\end{APACrefauthors}%
\unskip\
\newblock
\APACrefYearMonthDay{2008}{}{}.
\newblock
{\BBOQ}\APACrefatitle {Stimulus representation and the timing of
  reward-prediction errors in models of the dopamine system} {Stimulus
  representation and the timing of reward-prediction errors in models of the
  dopamine system}.{\BBCQ}
\newblock
\APACjournalVolNumPages{Neural computation}{20}{12}{3034--3054}.
\PrintBackRefs{\CurrentBib}

\bibitem [\protect \citeauthoryear {%
Ludvig%
, Sutton%
\BCBL {}\ \BBA {} Kehoe%
}{%
Ludvig%
\ \protect \BOthers {.}}{%
{\protect \APACyear {2012}}%
}]{%
ludvig2012evaluating}
\APACinsertmetastar {%
ludvig2012evaluating}%
\begin{APACrefauthors}%
Ludvig, E\BPBI A.%
, Sutton, R\BPBI S.%
\BCBL {}\ \BBA {} Kehoe, E\BPBI J.%
\end{APACrefauthors}%
\unskip\
\newblock
\APACrefYearMonthDay{2012}{}{}.
\newblock
{\BBOQ}\APACrefatitle {Evaluating the {TD} model of classical conditioning}
  {Evaluating the {TD} model of classical conditioning}.{\BBCQ}
\newblock
\APACjournalVolNumPages{Learning \& behavior}{40}{3}{305--319}.
\PrintBackRefs{\CurrentBib}

\bibitem [\protect \citeauthoryear {%
Ludvig%
, Sutton%
, Verbeek%
\BCBL {}\ \BBA {} Kehoe%
}{%
Ludvig%
\ \protect \BOthers {.}}{%
{\protect \APACyear {2009}}%
}]{%
ludvig2009computational}
\APACinsertmetastar {%
ludvig2009computational}%
\begin{APACrefauthors}%
Ludvig, E\BPBI A.%
, Sutton, R\BPBI S.%
, Verbeek, E.%
\BCBL {}\ \BBA {} Kehoe, E\BPBI J.%
\end{APACrefauthors}%
\unskip\
\newblock
\APACrefYearMonthDay{2009}{}{}.
\newblock
{\BBOQ}\APACrefatitle {A computational model of hippocampal function in trace
  conditioning} {A computational model of hippocampal function in trace
  conditioning}.{\BBCQ}
\newblock
\BIn{} \APACrefbtitle {Advances in Neural Information Processing Systems}
  {Advances in neural information processing systems}\ (\BPGS\ 993--1000).
\PrintBackRefs{\CurrentBib}

\bibitem [\protect \citeauthoryear {%
Luzardo%
}{%
Luzardo%
}{%
{\protect \APACyear {2018}}%
}]{%
luzardo2018rescorla}
\APACinsertmetastar {%
luzardo2018rescorla}%
\begin{APACrefauthors}%
Luzardo, A.%
\end{APACrefauthors}%
\unskip\
\newblock
\APACrefYear{2018}.
\unskip\
\newblock
\APACrefbtitle {The Rescorla-Wagner Drift-Diffusion Model} {The rescorla-wagner
  drift-diffusion model}\ \APACtypeAddressSchool {\BUPhD}{}{}.
\unskip\
\newblock
\APACaddressSchool {}{City, University of London}.
\PrintBackRefs{\CurrentBib}

\bibitem [\protect \citeauthoryear {%
Machado%
\ \protect \BOthers {.}}{%
Machado%
\ \protect \BOthers {.}}{%
{\protect \APACyear {2018}}%
}]{%
machado2018revisiting}
\APACinsertmetastar {%
machado2018revisiting}%
\begin{APACrefauthors}%
Machado, M\BPBI C.%
, Bellemare, M\BPBI G.%
, Talvitie, E.%
, Veness, J.%
, Hausknecht, M.%
\BCBL {}\ \BBA {} Bowling, M.%
\end{APACrefauthors}%
\unskip\
\newblock
\APACrefYearMonthDay{2018}{}{}.
\newblock
{\BBOQ}\APACrefatitle {Revisiting the arcade learning environment: Evaluation
  protocols and open problems for general agents} {Revisiting the arcade
  learning environment: Evaluation protocols and open problems for general
  agents}.{\BBCQ}
\newblock
\APACjournalVolNumPages{Journal of Artificial Intelligence
  Research}{61}{}{523--562}.
\PrintBackRefs{\CurrentBib}

\bibitem [\protect \citeauthoryear {%
Mackintosh%
}{%
Mackintosh%
}{%
{\protect \APACyear {1974}}%
}]{%
mackintosh1974psychology}
\APACinsertmetastar {%
mackintosh1974psychology}%
\begin{APACrefauthors}%
Mackintosh, N\BPBI J.%
\end{APACrefauthors}%
\unskip\
\newblock
\APACrefYear{1974}.
\newblock
\APACrefbtitle {The Psychology of Animal Learning.} {The psychology of animal
  learning.}
\newblock
\APACaddressPublisher{}{Academic Press}.
\PrintBackRefs{\CurrentBib}

\bibitem [\protect \citeauthoryear {%
Mahmood%
\ \BBA {} Sutton%
}{%
Mahmood%
\ \BBA {} Sutton%
}{%
{\protect \APACyear {2013}}%
}]{%
mahmood2013representation}
\APACinsertmetastar {%
mahmood2013representation}%
\begin{APACrefauthors}%
Mahmood, A\BPBI R.%
\BCBT {}\ \BBA {} Sutton, R\BPBI S.%
\end{APACrefauthors}%
\unskip\
\newblock
\APACrefYearMonthDay{2013}{}{}.
\newblock
{\BBOQ}\APACrefatitle {Representation search through generate and test}
  {Representation search through generate and test}.{\BBCQ}
\newblock
\BIn{} \APACrefbtitle {Workshops at the Twenty-Seventh Association for the
  Advancement of Artificial Intelligence Conference.} {Workshops at the
  twenty-seventh association for the advancement of artificial intelligence
  conference.}
\PrintBackRefs{\CurrentBib}

\bibitem [\protect \citeauthoryear {%
Menick%
\ \protect \BOthers {.}}{%
Menick%
\ \protect \BOthers {.}}{%
{\protect \APACyear {2020}}%
}]{%
menick2020practical}
\APACinsertmetastar {%
menick2020practical}%
\begin{APACrefauthors}%
Menick, J.%
, Elsen, E.%
, Evci, U.%
, Osindero, S.%
, Simonyan, K.%
\BCBL {}\ \BBA {} Graves, A.%
\end{APACrefauthors}%
\unskip\
\newblock
\APACrefYearMonthDay{2020}{}{}.
\newblock
{\BBOQ}\APACrefatitle {Practical Real Time Recurrent Learning with a Sparse
  Approximation} {Practical real time recurrent learning with a sparse
  approximation}.{\BBCQ}
\newblock
\BIn{} \APACrefbtitle {International Conference on Learning Representations.}
  {International conference on learning representations.}
\PrintBackRefs{\CurrentBib}

\bibitem [\protect \citeauthoryear {%
Modayil%
, White%
\BCBL {}\ \BBA {} Sutton%
}{%
Modayil%
\ \protect \BOthers {.}}{%
{\protect \APACyear {2014}}%
}]{%
modayil2014multi}
\APACinsertmetastar {%
modayil2014multi}%
\begin{APACrefauthors}%
Modayil, J.%
, White, A.%
\BCBL {}\ \BBA {} Sutton, R\BPBI S.%
\end{APACrefauthors}%
\unskip\
\newblock
\APACrefYearMonthDay{2014}{}{}.
\newblock
{\BBOQ}\APACrefatitle {Multi-timescale nexting in a reinforcement learning
  robot} {Multi-timescale nexting in a reinforcement learning robot}.{\BBCQ}
\newblock
\APACjournalVolNumPages{Adaptive Behavior}{22}{2}{146--160}.
\PrintBackRefs{\CurrentBib}

\bibitem [\protect \citeauthoryear {%
Mozer%
}{%
Mozer%
}{%
{\protect \APACyear {1989}}%
}]{%
mozer1989focused}
\APACinsertmetastar {%
mozer1989focused}%
\begin{APACrefauthors}%
Mozer, M\BPBI C.%
\end{APACrefauthors}%
\unskip\
\newblock
\APACrefYearMonthDay{1989}{}{}.
\newblock
{\BBOQ}\APACrefatitle {A focused back-propagation algorithm for temporal
  pattern recognition} {A focused back-propagation algorithm for temporal
  pattern recognition}.{\BBCQ}
\newblock
\APACjournalVolNumPages{Complex Systems}{3}{4}{349--381}.
\PrintBackRefs{\CurrentBib}

\bibitem [\protect \citeauthoryear {%
Nath%
\ \protect \BOthers {.}}{%
Nath%
\ \protect \BOthers {.}}{%
{\protect \APACyear {2019}}%
}]{%
nath2019training}
\APACinsertmetastar {%
nath2019training}%
\begin{APACrefauthors}%
Nath, S.%
, Liu, V.%
, Chan, A.%
, Li, X.%
, White, A.%
\BCBL {}\ \BBA {} White, M.%
\end{APACrefauthors}%
\unskip\
\newblock
\APACrefYearMonthDay{2019}{}{}.
\newblock
{\BBOQ}\APACrefatitle {Training recurrent neural networks online by learning
  explicit state variables} {Training recurrent neural networks online by
  learning explicit state variables}.{\BBCQ}
\newblock
\BIn{} \APACrefbtitle {International Conference on Learning Representations.}
  {International conference on learning representations.}
\PrintBackRefs{\CurrentBib}

\bibitem [\protect \citeauthoryear {%
Osband%
\ \protect \BOthers {.}}{%
Osband%
\ \protect \BOthers {.}}{%
{\protect \APACyear {2020}}%
}]{%
osband2019behaviour}
\APACinsertmetastar {%
osband2019behaviour}%
\begin{APACrefauthors}%
Osband, I.%
, Doron, Y.%
, Hessel, M.%
, Aslanides, J.%
, Sezener, E.%
, Saraiva, A.%
\BDBL {}Hasselt, H\BPBI V.%
\end{APACrefauthors}%
\unskip\
\newblock
\APACrefYearMonthDay{2020}{}{}.
\newblock
{\BBOQ}\APACrefatitle {Behaviour Suite for Reinforcement Learning} {Behaviour
  suite for reinforcement learning}.{\BBCQ}
\newblock
\BIn{} \APACrefbtitle {International Conference on Learning Representations.}
  {International conference on learning representations.}
\PrintBackRefs{\CurrentBib}

\bibitem [\protect \citeauthoryear {%
Osband%
, Van~Roy%
, Russo%
\BCBL {}\ \BBA {} Wen%
}{%
Osband%
\ \protect \BOthers {.}}{%
{\protect \APACyear {2019}}%
}]{%
osband2019deep}
\APACinsertmetastar {%
osband2019deep}%
\begin{APACrefauthors}%
Osband, I.%
, Van~Roy, B.%
, Russo, D\BPBI J.%
\BCBL {}\ \BBA {} Wen, Z.%
\end{APACrefauthors}%
\unskip\
\newblock
\APACrefYearMonthDay{2019}{}{}.
\newblock
{\BBOQ}\APACrefatitle {Deep Exploration via Randomized Value Functions.} {Deep
  exploration via randomized value functions.}{\BBCQ}
\newblock
\APACjournalVolNumPages{Journal of Machine Learning Research}{20}{124}{1--62}.
\PrintBackRefs{\CurrentBib}

\bibitem [\protect \citeauthoryear {%
Parisotto%
\ \BBA {} Salakhutdinov%
}{%
Parisotto%
\ \BBA {} Salakhutdinov%
}{%
{\protect \APACyear {2021}}%
}]{%
parisotto2021efficient}
\APACinsertmetastar {%
parisotto2021efficient}%
\begin{APACrefauthors}%
Parisotto, E.%
\BCBT {}\ \BBA {} Salakhutdinov, R.%
\end{APACrefauthors}%
\unskip\
\newblock
\APACrefYearMonthDay{2021}{}{}.
\newblock
{\BBOQ}\APACrefatitle {Efficient Transformers in Reinforcement Learning using
  Actor-Learner Distillation} {Efficient transformers in reinforcement learning
  using actor-learner distillation}.{\BBCQ}
\newblock
\BIn{} \APACrefbtitle {International Conference on Learning Representations.}
  {International conference on learning representations.}
\PrintBackRefs{\CurrentBib}

\bibitem [\protect \citeauthoryear {%
Parisotto%
\ \protect \BOthers {.}}{%
Parisotto%
\ \protect \BOthers {.}}{%
{\protect \APACyear {2020}}%
}]{%
parisotto2019stabilizing}
\APACinsertmetastar {%
parisotto2019stabilizing}%
\begin{APACrefauthors}%
Parisotto, E.%
, Song, F.%
, Rae, J.%
, Pascanu, R.%
, Gulcehre, C.%
, Jayakumar, S.%
\BDBL {}Hadsell, R.%
\end{APACrefauthors}%
\unskip\
\newblock
\APACrefYearMonthDay{2020}{13--18 Jul}{}.
\newblock
{\BBOQ}\APACrefatitle {Stabilizing Transformers for Reinforcement Learning}
  {Stabilizing transformers for reinforcement learning}.{\BBCQ}
\newblock
\BIn{} \APACrefbtitle {Proceedings of the 37th International Conference on
  Machine Learning} {Proceedings of the 37th international conference on
  machine learning}\ (\BVOL~119, \BPGS\ 7487--7498).
\newblock
\APACaddressPublisher{}{PMLR}.
\PrintBackRefs{\CurrentBib}

\bibitem [\protect \citeauthoryear {%
Pavlov%
}{%
Pavlov%
}{%
{\protect \APACyear {1927}}%
}]{%
pavlov1927conditioned}
\APACinsertmetastar {%
pavlov1927conditioned}%
\begin{APACrefauthors}%
Pavlov, I\BPBI P.%
\end{APACrefauthors}%
\unskip\
\newblock
\APACrefYear{1927}.
\newblock
\APACrefbtitle {Conditioned reflexes: an investigation of the physiological
  activity of the cerebral cortex} {Conditioned reflexes: an investigation of
  the physiological activity of the cerebral cortex}\ (\BVOL~3).
\newblock
\APACaddressPublisher{}{london: oxford University Press}.
\PrintBackRefs{\CurrentBib}

\bibitem [\protect \citeauthoryear {%
Rivest%
, Kalaska%
\BCBL {}\ \BBA {} Bengio%
}{%
Rivest%
\ \protect \BOthers {.}}{%
{\protect \APACyear {2014}}%
}]{%
rivest2014conditioning}
\APACinsertmetastar {%
rivest2014conditioning}%
\begin{APACrefauthors}%
Rivest, F.%
, Kalaska, J\BPBI F.%
\BCBL {}\ \BBA {} Bengio, Y.%
\end{APACrefauthors}%
\unskip\
\newblock
\APACrefYearMonthDay{2014}{}{}.
\newblock
{\BBOQ}\APACrefatitle {Conditioning and time representation in long short-term
  memory networks} {Conditioning and time representation in long short-term
  memory networks}.{\BBCQ}
\newblock
\APACjournalVolNumPages{Biological Cybernetics}{108}{1}{23--48}.
\PrintBackRefs{\CurrentBib}

\bibitem [\protect \citeauthoryear {%
Schneiderman%
}{%
Schneiderman%
}{%
{\protect \APACyear {1966}}%
}]{%
schneiderman1966interstimulus}
\APACinsertmetastar {%
schneiderman1966interstimulus}%
\begin{APACrefauthors}%
Schneiderman, N.%
\end{APACrefauthors}%
\unskip\
\newblock
\APACrefYearMonthDay{1966}{}{}.
\newblock
{\BBOQ}\APACrefatitle {Interstimulus interval function of the nictitating
  membrane response of the rabbit under delay versus trace conditioning.}
  {Interstimulus interval function of the nictitating membrane response of the
  rabbit under delay versus trace conditioning.}{\BBCQ}
\newblock
\APACjournalVolNumPages{Journal of Comparative and Physiological
  Psychology}{62}{3}{397}.
\PrintBackRefs{\CurrentBib}

\bibitem [\protect \citeauthoryear {%
Sutton%
}{%
Sutton%
}{%
{\protect \APACyear {1988}}%
}]{%
sutton1988learning}
\APACinsertmetastar {%
sutton1988learning}%
\begin{APACrefauthors}%
Sutton, R\BPBI S.%
\end{APACrefauthors}%
\unskip\
\newblock
\APACrefYearMonthDay{1988}{}{}.
\newblock
{\BBOQ}\APACrefatitle {Learning to predict by the methods of temporal
  differences} {Learning to predict by the methods of temporal
  differences}.{\BBCQ}
\newblock
\APACjournalVolNumPages{Machine Learning}{3}{1}{9--44}.
\PrintBackRefs{\CurrentBib}

\bibitem [\protect \citeauthoryear {%
Sutton%
}{%
Sutton%
}{%
{\protect \APACyear {1992}}%
}]{%
sutton1992adapting}
\APACinsertmetastar {%
sutton1992adapting}%
\begin{APACrefauthors}%
Sutton, R\BPBI S.%
\end{APACrefauthors}%
\unskip\
\newblock
\APACrefYearMonthDay{1992}{}{}.
\newblock
{\BBOQ}\APACrefatitle {Adapting bias by gradient descent: An incremental
  version of delta-bar-delta} {Adapting bias by gradient descent: An
  incremental version of delta-bar-delta}.{\BBCQ}
\newblock
\BIn{} \APACrefbtitle {Proceedings of the Tenth National Conference on
  Artificial Intelligence} {Proceedings of the tenth national conference on
  artificial intelligence}\ (\BPGS\ 171--176).
\PrintBackRefs{\CurrentBib}

\bibitem [\protect \citeauthoryear {%
Sutton%
\ \BBA {} Barto%
}{%
Sutton%
\ \BBA {} Barto%
}{%
{\protect \APACyear {1990}}%
}]{%
sutton1990time}
\APACinsertmetastar {%
sutton1990time}%
\begin{APACrefauthors}%
Sutton, R\BPBI S.%
\BCBT {}\ \BBA {} Barto, A\BPBI G.%
\end{APACrefauthors}%
\unskip\
\newblock
\APACrefYearMonthDay{1990}{}{}.
\newblock
{\BBOQ}\APACrefatitle {Time-derivative models of pavlovian reinforcement. {In
  M. Gabriel and J. Moore (Eds.)}} {Time-derivative models of pavlovian
  reinforcement. {In M. Gabriel and J. Moore (Eds.)}}.{\BBCQ}
\newblock
\APACjournalVolNumPages{Learning and Computational Neuroscience: Foundations of
  Adaptive Networks}{}{}{497--537}.
\PrintBackRefs{\CurrentBib}

\bibitem [\protect \citeauthoryear {%
Sutton%
\ \BBA {} Barto%
}{%
Sutton%
\ \BBA {} Barto%
}{%
{\protect \APACyear {2018}}%
}]{%
sutton2018reinforcement}
\APACinsertmetastar {%
sutton2018reinforcement}%
\begin{APACrefauthors}%
Sutton, R\BPBI S.%
\BCBT {}\ \BBA {} Barto, A\BPBI G.%
\end{APACrefauthors}%
\unskip\
\newblock
\APACrefYear{2018}.
\newblock
\APACrefbtitle {Reinforcement learning: An introduction} {Reinforcement
  learning: An introduction}.
\newblock
\APACaddressPublisher{}{MIT press}.
\PrintBackRefs{\CurrentBib}

\bibitem [\protect \citeauthoryear {%
Sutton%
, Koop%
\BCBL {}\ \BBA {} Silver%
}{%
Sutton%
\ \protect \BOthers {.}}{%
{\protect \APACyear {2007}}%
}]{%
sutton2007role}
\APACinsertmetastar {%
sutton2007role}%
\begin{APACrefauthors}%
Sutton, R\BPBI S.%
, Koop, A.%
\BCBL {}\ \BBA {} Silver, D.%
\end{APACrefauthors}%
\unskip\
\newblock
\APACrefYearMonthDay{2007}{}{}.
\newblock
{\BBOQ}\APACrefatitle {On the role of tracking in stationary environments} {On
  the role of tracking in stationary environments}.{\BBCQ}
\newblock
\BIn{} \APACrefbtitle {Proceedings of the 24th International Conference on
  Machine Learning} {Proceedings of the 24th international conference on
  machine learning}\ (\BPGS\ 871--878).
\PrintBackRefs{\CurrentBib}

\bibitem [\protect \citeauthoryear {%
Sutton%
\ \BBA {} Whitehead%
}{%
Sutton%
\ \BBA {} Whitehead%
}{%
{\protect \APACyear {1993}}%
}]{%
sutton2014online}
\APACinsertmetastar {%
sutton2014online}%
\begin{APACrefauthors}%
Sutton, R\BPBI S.%
\BCBT {}\ \BBA {} Whitehead, S\BPBI D.%
\end{APACrefauthors}%
\unskip\
\newblock
\APACrefYearMonthDay{1993}{}{}.
\newblock
{\BBOQ}\APACrefatitle {Online learning with random representations} {Online
  learning with random representations}.{\BBCQ}
\newblock
\BIn{} \APACrefbtitle {Proceedings of the Tenth International Conference on
  Machine Learning} {Proceedings of the tenth international conference on
  machine learning}\ (\BPGS\ 314--321).
\PrintBackRefs{\CurrentBib}

\bibitem [\protect \citeauthoryear {%
Tallec%
\ \BBA {} Ollivier%
}{%
Tallec%
\ \BBA {} Ollivier%
}{%
{\protect \APACyear {2018}}%
}]{%
tallec2017unbiased}
\APACinsertmetastar {%
tallec2017unbiased}%
\begin{APACrefauthors}%
Tallec, C.%
\BCBT {}\ \BBA {} Ollivier, Y.%
\end{APACrefauthors}%
\unskip\
\newblock
\APACrefYearMonthDay{2018}{}{}.
\newblock
{\BBOQ}\APACrefatitle {Unbiased Online Recurrent Optimization} {Unbiased online
  recurrent optimization}.{\BBCQ}
\newblock
\BIn{} \APACrefbtitle {International Conference on Learning Representations.}
  {International conference on learning representations.}
\PrintBackRefs{\CurrentBib}

\bibitem [\protect \citeauthoryear {%
Todorov%
, Erez%
\BCBL {}\ \BBA {} Tassa%
}{%
Todorov%
\ \protect \BOthers {.}}{%
{\protect \APACyear {2012}}%
}]{%
todorov2012mujoco}
\APACinsertmetastar {%
todorov2012mujoco}%
\begin{APACrefauthors}%
Todorov, E.%
, Erez, T.%
\BCBL {}\ \BBA {} Tassa, Y.%
\end{APACrefauthors}%
\unskip\
\newblock
\APACrefYearMonthDay{2012}{}{}.
\newblock
{\BBOQ}\APACrefatitle {Mujoco: A physics engine for model-based control}
  {Mujoco: A physics engine for model-based control}.{\BBCQ}
\newblock
\BIn{} \APACrefbtitle {2012 IEEE/RSJ International Conference on Intelligent
  Robots and Systems} {2012 ieee/rsj international conference on intelligent
  robots and systems}\ (\BPGS\ 5026--5033).
\PrintBackRefs{\CurrentBib}

\bibitem [\protect \citeauthoryear {%
Tucker%
\ \protect \BOthers {.}}{%
Tucker%
\ \protect \BOthers {.}}{%
{\protect \APACyear {2018}}%
}]{%
tucker2018mirage}
\APACinsertmetastar {%
tucker2018mirage}%
\begin{APACrefauthors}%
Tucker, G.%
, Bhupatiraju, S.%
, Gu, S.%
, Turner, R.%
, Ghahramani, Z.%
\BCBL {}\ \BBA {} Levine, S.%
\end{APACrefauthors}%
\unskip\
\newblock
\APACrefYearMonthDay{2018}{}{}.
\newblock
{\BBOQ}\APACrefatitle {The mirage of action-dependent baselines in
  reinforcement learning} {The mirage of action-dependent baselines in
  reinforcement learning}.{\BBCQ}
\newblock
\BIn{} \APACrefbtitle {International Conference on Machine Learning}
  {International conference on machine learning}\ (\BPGS\ 5015--5024).
\PrintBackRefs{\CurrentBib}

\bibitem [\protect \citeauthoryear {%
van Hasselt%
\ \BBA {} Sutton%
}{%
van Hasselt%
\ \BBA {} Sutton%
}{%
{\protect \APACyear {2015}}%
}]{%
van2015learning}
\APACinsertmetastar {%
van2015learning}%
\begin{APACrefauthors}%
van Hasselt, H.%
\BCBT {}\ \BBA {} Sutton, R\BPBI S.%
\end{APACrefauthors}%
\unskip\
\newblock
\APACrefYearMonthDay{2015}{}{}.
\newblock
{\BBOQ}\APACrefatitle {Learning to predict independent of span} {Learning to
  predict independent of span}.{\BBCQ}
\newblock
\APACjournalVolNumPages{arXiv preprint arXiv:1508.04582}{}{}{}.
\PrintBackRefs{\CurrentBib}

\bibitem [\protect \citeauthoryear {%
Wagner%
}{%
Wagner%
}{%
{\protect \APACyear {1978}}%
}]{%
wagner1978expectancies}
\APACinsertmetastar {%
wagner1978expectancies}%
\begin{APACrefauthors}%
Wagner, A\BPBI R.%
\end{APACrefauthors}%
\unskip\
\newblock
\APACrefYearMonthDay{1978}{}{}.
\newblock
{\BBOQ}\APACrefatitle {Expectancies and the priming of STM} {Expectancies and
  the priming of stm}.{\BBCQ}
\newblock
\APACjournalVolNumPages{Cognitive Processes in Animal Behavior}{}{}{177--209}.
\PrintBackRefs{\CurrentBib}

\bibitem [\protect \citeauthoryear {%
Wayne%
\ \protect \BOthers {.}}{%
Wayne%
\ \protect \BOthers {.}}{%
{\protect \APACyear {2018}}%
}]{%
wayne2018unsupervised}
\APACinsertmetastar {%
wayne2018unsupervised}%
\begin{APACrefauthors}%
Wayne, G.%
, Hung, C.%
, Amos, D.%
, Mirza, M.%
, Ahuja, A.%
, Grabska-Barwinska, A.%
\BDBL {}Lillicrap, T.%
\end{APACrefauthors}%
\unskip\
\newblock
\APACrefYearMonthDay{2018}{}{}.
\newblock
{\BBOQ}\APACrefatitle {Unsupervised predictive memory in a goal-directed agent}
  {Unsupervised predictive memory in a goal-directed agent}.{\BBCQ}
\newblock
\APACjournalVolNumPages{arXiv preprint arXiv:1803.10760}{}{}{}.
\PrintBackRefs{\CurrentBib}

\bibitem [\protect \citeauthoryear {%
Whiteson%
, Tanner%
\BCBL {}\ \BBA {} White%
}{%
Whiteson%
\ \protect \BOthers {.}}{%
{\protect \APACyear {2010}}%
}]{%
whiteson2010report}
\APACinsertmetastar {%
whiteson2010report}%
\begin{APACrefauthors}%
Whiteson, S.%
, Tanner, B.%
\BCBL {}\ \BBA {} White, A.%
\end{APACrefauthors}%
\unskip\
\newblock
\APACrefYearMonthDay{2010}{}{}.
\newblock
{\BBOQ}\APACrefatitle {Report on the 2008 reinforcement learning competition}
  {Report on the 2008 reinforcement learning competition}.{\BBCQ}
\newblock
\APACjournalVolNumPages{AI Magazine}{31}{2}{81--81}.
\PrintBackRefs{\CurrentBib}

\bibitem [\protect \citeauthoryear {%
D\BPBI A.~Williams%
, Todd%
, Chubala%
\BCBL {}\ \BBA {} Ludvig%
}{%
D\BPBI A.~Williams%
\ \protect \BOthers {.}}{%
{\protect \APACyear {2017}}%
}]{%
williams2017intertrial}
\APACinsertmetastar {%
williams2017intertrial}%
\begin{APACrefauthors}%
Williams, D\BPBI A.%
, Todd, T\BPBI P.%
, Chubala, C\BPBI M.%
\BCBL {}\ \BBA {} Ludvig, E\BPBI A.%
\end{APACrefauthors}%
\unskip\
\newblock
\APACrefYearMonthDay{2017}{}{}.
\newblock
{\BBOQ}\APACrefatitle {Intertrial unconditioned stimuli differentially impact
  trace conditioning} {Intertrial unconditioned stimuli differentially impact
  trace conditioning}.{\BBCQ}
\newblock
\APACjournalVolNumPages{Learning \& Behavior}{45}{1}{49--61}.
\PrintBackRefs{\CurrentBib}

\bibitem [\protect \citeauthoryear {%
R\BPBI J.~Williams%
\ \BBA {} Peng%
}{%
R\BPBI J.~Williams%
\ \BBA {} Peng%
}{%
{\protect \APACyear {1990}}%
}]{%
williams1990efficient}
\APACinsertmetastar {%
williams1990efficient}%
\begin{APACrefauthors}%
Williams, R\BPBI J.%
\BCBT {}\ \BBA {} Peng, J.%
\end{APACrefauthors}%
\unskip\
\newblock
\APACrefYearMonthDay{1990}{}{}.
\newblock
{\BBOQ}\APACrefatitle {An efficient gradient-based algorithm for on-line
  training of recurrent network trajectories} {An efficient gradient-based
  algorithm for on-line training of recurrent network trajectories}.{\BBCQ}
\newblock
\APACjournalVolNumPages{Neural Computation}{2}{4}{490--501}.
\PrintBackRefs{\CurrentBib}

\bibitem [\protect \citeauthoryear {%
R\BPBI J.~Williams%
\ \BBA {} Zipser%
}{%
R\BPBI J.~Williams%
\ \BBA {} Zipser%
}{%
{\protect \APACyear {1989}}%
}]{%
williams1989learning}
\APACinsertmetastar {%
williams1989learning}%
\begin{APACrefauthors}%
Williams, R\BPBI J.%
\BCBT {}\ \BBA {} Zipser, D.%
\end{APACrefauthors}%
\unskip\
\newblock
\APACrefYearMonthDay{1989}{}{}.
\newblock
{\BBOQ}\APACrefatitle {A learning algorithm for continually running fully
  recurrent neural networks} {A learning algorithm for continually running
  fully recurrent neural networks}.{\BBCQ}
\newblock
\APACjournalVolNumPages{Neural Computation}{1}{2}{270--280}.
\PrintBackRefs{\CurrentBib}

\bibitem [\protect \citeauthoryear {%
Zhang%
\ \BBA {} Sutton%
}{%
Zhang%
\ \BBA {} Sutton%
}{%
{\protect \APACyear {2017}}%
}]{%
zhang2017deeper}
\APACinsertmetastar {%
zhang2017deeper}%
\begin{APACrefauthors}%
Zhang, S.%
\BCBT {}\ \BBA {} Sutton, R\BPBI S.%
\end{APACrefauthors}%
\unskip\
\newblock
\APACrefYearMonthDay{2017}{}{}.
\newblock
{\BBOQ}\APACrefatitle {A deeper look at experience replay} {A deeper look at
  experience replay}.{\BBCQ}
\newblock
\APACjournalVolNumPages{arXiv preprint arXiv:1712.01275}{}{}{}.
\PrintBackRefs{\CurrentBib}

\end{thebibliography}
\end{document}